# Information-driven Path Planning for Hybrid Aerial Underwater Vehicles

Zheng Zeng[1,2], Chengke Xiong[1,2], Xinyi Yuan[1], Yulin Bai[1], Yufei Jin[1], Di Lu[1,2], Lian Lian[1,2]

[1] School of Oceanography, Shanghai Jiao Tong University, Shanghai, China.

[2] State Key Laboratory of Ocean Engineering, Shanghai Jiao Tong University, Shanghai, China.

zheng.zeng@sjtu.edu.cn; xiongchengke@sjtu.edu.cn

**Abstrac**t: This paper presents a novel Rapidly-exploring Adaptive Sampling Tree (RAST) algorithm for the adaptive sampling mission of a hybrid aerial underwater vehicle (HAUV) in an air-sea 3D environment. This algorithm innovatively combines the tournament-based point selection sampling strategy, the information heuristic search process and the framework of Rapidly-exploring Random Tree (RRT) algorithm. Hence can guide the vehicle to the region of interest to scientists for sampling and generate a collision-free path for maximizing information collection by the HAUV under the constraints of environmental effects of currents or wind and limited budget. The simulation results show that the fast search adaptive sampling tree algorithm has higher optimization performance, faster solution speed and better stability than the Rapidly-exploring Information Gathering Tree (RIGT) algorithm and the particle swarm optimization (PSO) algorithm.

1. INTRODUCTION

Efficient observation of hydrometeorological parameters at the sea-air interface to obtain high-quality, high-resolution data could provide accurate boundary conditions for numerical prediction models of the ocean and atmosphere, which is essential for the study of physical mechanisms of sea-air interactions, accurate forecasting of typhoons (hurricanes), and marine disaster prevention and mitigation [1]. For the sea-air adaptive sampling mission, the observation platforms are desired to carry various sensors to acquire the ocean physical, chemical and biological measurement data intelligently and autonomously in real-time. Recently developed observation platforms including Unmanned Aerial Vehicles (UAVs) [2], Unmanned Surface Vehicles (USVs) [3], Autonomous Underwater Vehicles (AUVs) [4], Underwater Gliders (UGs) [5], and other unmanned vehicles [6] are increasingly used by marine scientists for adaptive ocean observation and sampling. UAVs fly fast and can capture and measure changing atmospheric phenomena in the air. USVs sail fast, observe oceanic phenomena at the surface and can carry small AUVs to a designated location for deployment if needed. AUVs are mobile when fully submersed, capable of observing and sampling small-scale oceanic phenomena underwater. Underwater gliders can work continuously underwater for several months to observe large and medium scale ocean phenomena. However, the above-mentioned mobile observation platforms cannot simultaneously conduct joint sea-air observations of oceanic and atmospheric phenomena with 3D distribution and high temporal and spatial variability in specific sea areas.

In recent years, a class of highly mobile Hybrid Aerial Underwater Vehicle (HAUV) [7, 8], which can conduct air, surface and underwater surveys, has come into being. The HAUVs can carry air-sea optical observers and physicochemical sensors for air, surface, and underwater detection and data acquisition [9]. Compared with mobile observation platforms that can only operate in a single specific environmental medium, HAUVs have two modes of operation: aerial flight and underwater diving, so they can switch their modes of operation independently according to environmental information and mission requirements[10]; in addition, HAUVs have the advantages of higher mobility and lower operating costs[11] and can perform continuous, high-quality, and high-precision ocean and atmospheric characteristic parameters in the air-sea 3D environment. This reduces the total cost of sea-air stereo observation and sampling and improves the operational efficiency and the amount of actual data acquisition in a single mission.

The use of HAUVs to perform sea-air stereo adaptive observation and sampling tasks can realize the simultaneous real-time observation of hydrometeorological data at the sea-air interface and provide a new observation means for multi-scale sea-air interaction research as shown in Figure 1. An HAUV could be launched from shore or a surface vehicle where upon a path planning system could be used to generate a trajectory that lead the vehicle to a work site, perform a survey, and then return to shore or deck completely on its own. The surface vehicle will provide the HAUV with localization support, acoustic localization was used to determine the HAUV position underwater. The Blueprint SeaTrac miniature USBL could be integrated into both HAUV and surface vehicle for simultaneous localization and data exchange[12]. To fulfill this mission, it is necessary for the path planner to consider the motion characteristics, the energy consumption of operation, and the time constraints of the HAUV in both ocean and atmospheric. Furthermore, due to the cross-field nature of sea-air interface observation, it is necessary to combine the definition and analysis methods of both ocean and atmospheric phenomena to determine the type, location, and time of the most exploitable hydrometeorological parameters in a sea area and plan feasible and optimal sampling paths [4]. In practical mission scenarios, joint sea-air stereo observation and adaptive sampling are directly associated with the path planning system of the vehicle. The path planning systems should determine the target observation area accessible by the HAUV based on the sea-air environment model and guide the HAUV to navigate along the optimal sampling path to collect the parameter data desired by marine scientists.

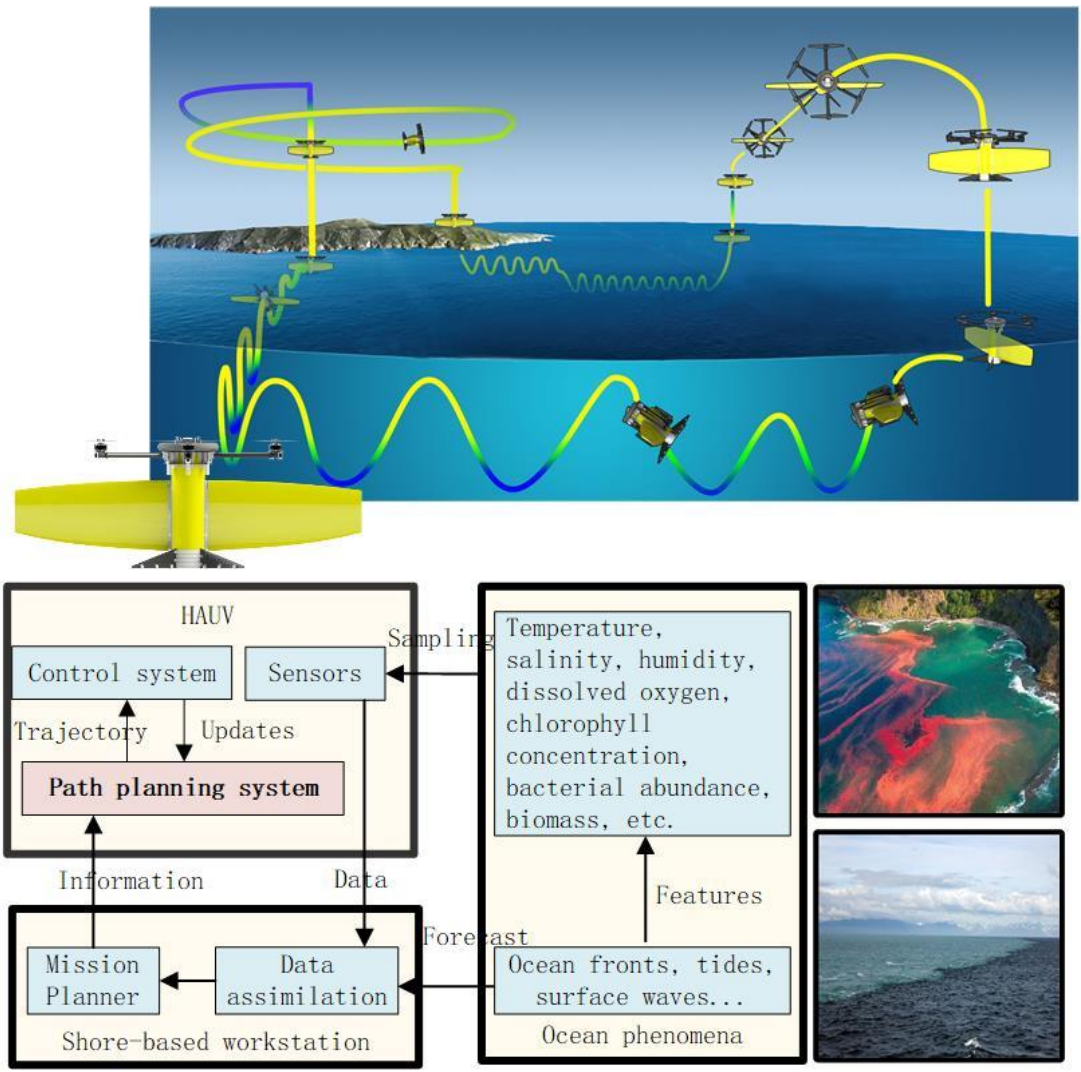

Figure 1 Flow chart of HAUV for ocean-atmospheric adaptive sampling

Therefore, the performance of the path planning system is an important manifestation of the intelligence level of the HAUV for air-sea stereo observation, which determines the sampling efficiency and observation capability of the HAUV in the joint air-sea observation mission. The existing path planning research mainly focuses on the information-driven path planning for unmanned vehicles operating in a single medium, air or underwater. There is an urgent need to develop a path planning system applicable to the sea-air 3D adaptive sampling of HAUVs. It should combine complex sea-air information with the motion performance and system constraints of HAUVs in a specific medium and adopt suitable path optimization algorithms so that the generated sampling paths can guide the HAUVs to the areas with rich environmental information for sampling.

Existing information-driven path planning algorithms include branch-and-bound method, random sampling algorithm, population intelligence optimization algorithm, etc. The random sampling algorithm and the classical RRT* algorithm have low computational complexity and are suitable for solving path planning problems in high-dimensional space. It can guarantee the probabilistic completeness and asymptotic optimality of the solution. A review of existing information-driven path planning algorithms is presented in Section 2. This paper will improve and expand on the RRT* algorithm by innovatively integrating the sampling strategy based on the tournament point selection method, the information heuristic search process and the framework of the RRT* algorithm to design an algorithm applicable to information-driven path planning for HAUVs.

The main contributions of this work are listed as follows:

- We formulate the information-driven HAUV path planning problem under the constraints of environmental effects of currents or wind and limited budget.
- We present a novel RAST path planning method for HAUV that can guide the vehicle to the region of interest for sampling and generate a collision-free path for maximizing information collection.
- We compared the important optimization techniques applied to HAUV path planning in several scenarios, the weaknesses and strengths of each optimization technique have been stated.

The rest of this paper is organized as follows. In Section 3, the information-driven path planning problem for a HAUV is formulated. In Section 4, we propose the RAST* algorithm and design four different forms of the RAST* algorithm based on the theory of this algorithm and intend to investigate which optimization method can improve the computational speed and solution accuracy of the RAST* algorithm by introducing comparative experiments. In Section 5, the classical RIGT algorithm and PSO algorithm are used as comparison algorithms, and the performance of different sampling path optimization algorithms is compared through simulation experiments. The prospect of applying RAST* algorithm in different scenarios is discussed. Concluding remarks are then presented in Section 6.

2. RELATED WORK

Information-driven path planning is one of the key technologies for adaptive sampling of unmanned aerial vehicles (UAVs). It aims to generate sampling paths that allow the UAV, within the constraints of a limited budget (e.g., energy, mission time, etc.), to maximize the amount of information observed and collected in the target area [13-15]. Commonly used information-driven path planning algorithms include branch-and-bound, population intelligence optimization algorithms and random sampling algorithms.

2.1 Information-driven path planning based on branch delimitation algorithm

The branch-and-bound method is widely used for solving constrained optimization problems and can efficiently search a finite number of feasible solution spaces systematically [16]. Namik *et al.* proposes using the branch-and-bound method to solve the optimal sampling path for a single AUV and multiple AUVs to maximize the sum of line integrals of the uncertainty values along the entire path [17]. Amarjeet *et al.* investigated the use of unmanned vehicles for sampling tasks with effective spatial coverage in application scenarios that require effective monitoring of spatio-temporal dynamic environments, such as water quality monitoring in rivers and lakes [18]. Jonathan *et al.* introduced a pilot measurement procedure in the branch-and-bound method to solve the adaptive ocean sampling path planning problem of AUV[19]. The authors verified that the branch-and-bound method with pilot measurement is more efficient than that without pilot measurement. Paul *et al.* explored using AUVs with multiple sensors to build water quality models to help assess important watershed environmental hazards [20]. The authors propose two information-driven path planning algorithms, branch-and-bound and cross-entropy optimization, to select the future sampling locations of the AUV under the condition of the kinematic constraints of the AUV. The effectiveness of the proposed method is verified by simulation and field experiments. The branch-and-bound method is simple in structure and fast in solving and is suitable for information-driven path planning problems in small-scale static environments.

2.2 Information-driven path planning based on population intelligence optimization algorithm

Population intelligence optimization algorithms such as genetic algorithms (GA), PSO algorithms, and ant colony optimization (ACO) have been applied to information-driven path planning problems in the literature. Kevin *et al.* used genetic algorithms to plan adaptive sampling paths for multiple underwater gliders [21, 22]. Mario *et al.* uses GA to search for a water-free path that maximizes the sweeping area of the lake by a manned boat to monitor the lake environment[23]. Sergey *et al.* proposed a fully nonlinear GA for solving the optimal sampling path of AUV. The authors verified the optimization capability of GA by comparing it with the mower method and A* algorithm through simulation experiments [24]. Hexiong *et al.* integrates the fuzzy integrated evaluation method into a multi-objective PSO algorithm to solve the path planning problem of adaptive sampling with multiple AUVs in a dynamic ocean environment. This method uses sampling value and energy consumption as multi-objective cost functions [25]. Chengke *et al.* proposed an elite group-based PSO algorithm for planning AUV paths to maximize marine environmental feature information collection in a static ocean environment [26]. Giancarlo *et al.* Colmenares used an ACO algorithm to solve a single unmanned vehicle data collection task planning problem, planning a path for the vehicle to maximize the amount of water quality sampled [27]. Chengke *et al.* introduced the Delaunay space partitioning strategy into an ACO to form a path planning system that can effectively guide the path planning system can effectively guide the vehicle to the area of interest to the scientists [28]. Yichen *et al.* used an ACO algorithm to plan the sampling path of AUV to maximize the acquisition of temperature data in the 3D environment of temperature distribution of the regional ocean model system [29]. However, only the simulation experiments of AUV in a 3D unobstructed and current-free ocean environment model were conducted. Other population intelligence optimization algorithms, including simulated annealing algorithm [30, 31], differential evolution algorithm [32], covariance matrix adaptive evolution algorithm [33, 34], etc. have also been applied to information-driven path planning problems in the literature. The population intelligence optimization algorithm uses a population search model, simple theory, and easy application; however, as the dimensionality and size of the search space increases, the convergence speed decreases sharply, and it is easy to fall into local optimal solutions.

2.3 Information-driven path planning based on the random sampling algorithm

One of the random sampling algorithms widely used in path planning is the Rapidly-exploring Random Tree algorithm (RRT)[35]. Geoffrey *et al.* proposed a fast search information gathering tree algorithm based on the RRT* algorithm. It can plan paths that maximize information collection for AUVs with pre-defined constraints (e.g., energy or time constraints). It is also proved that the paths obtained after optimization are asymptotically optimal [36]. Subsequently, Maani *et al.* proposed an incremental search information collection tree algorithm based on this algorithm that can compute the sampling paths of USVs online [37]. Rongxin *et al.* proposed a multidimensional fast search random tree algorithm based on mutual information for solving multiple AUVs to maximize the understanding of the region of interest while minimizing the estimation error of the optimal sampling path. The authors verified the feasibility and effectiveness of the multidimensional fast search random tree algorithm based on mutual information through pool experiments [38]. Alberto *et al.* proposed a two-step path planning strategy for robots to collect information about unknown physical processes efficiently. The RRT algorithm is used to determine the location points not yet visited by the robot in the first step and plan a sampling path that maximizes information collection while minimizing the path cost in the second step [39]. Chengke *et al.* incorporated a tournament selection method into the RRT* algorithm and proposed an adaptive sampling-based path planning system to generate an unmanned vehicle sampling path that maximizes information collection under the influence of obstacle environments and sea currents [40]. The RRT* algorithm has low computational complexity, which grows slowly as the size of the space increases. It is suitable for solving information-driven path planning problems in a large high-dimensional space while guaranteeing the solution's probabilistic completeness and asymptotic optimality. It can ensure the probabilistic completeness and asymptotic optimality of the solution.

Joint air-sea information-driven path planning is a high-dimensional, multi-constraint optimization problem. The computational effort and complexity of the problem will increase exponentially as the search space range increases, so the information-driven path planning module for HAUVs needs to adopt an optimization algorithm that can quickly solve high-dimensional complex problems. Synthesizing the current research status of the above three major classes of algorithms, as shown in Table 1.

TABLE 1 Parameter settings for the HAUV and all algorithms

| Algorithm | Advantages | Disadvantages | Applicability |
|---|---|---|---|
| Branch-and-bound method | Simple structure, fast solution speed | Linear, discrete search | Suitable for small scale Static environment |
| Group intelligent optimization algorithm | Simple theory, easy to apply | Computation speed is greatly affected by space scale, easy to fall into the local optimum solution | Suitable for medium and small scale Small and medium scale environment |
| Random sampling algorithm | Low computational complexity, low speed of computation affected by spatial scale | Asymptotic optimization | Suitable for large scale High-dimensional environment |

## 3. INFORMATION-DRIVEN PATH PLANNING PROBLEM FOR HAUVS IN A 3D ENVIRONMENT IN AIR AND SEA

The goal of the global information-driven path planning system for HAUVs is to find a globally optimal sampling path $\mathbb{P}^*$ from the set of feasible paths $\Psi_{\mathbb{P}}$ that efficiently avoid obstacles $C_{obs}$ (e.g., ships, reefs, islands, etc.) and maximize the observation and collection of characteristic information (e.g., seawater temperature, salinity, chlorophyll fluorescence, dissolved oxygen concentration, air temperature, pressure, carbon dioxide fluxes, turbulent heat fluxes, etc.) of interest to scientists. The impact of the wind and flow fields $V_c$ on the vehicle should be fully considered.

Before discussing the problems studied in this paper, the following assumptions are made.

*Assumptions 1*: this paper primarily focuses on a high-level planning architecture with simplified dynamics enabling it to find the optimum trajectory for maximizing information collection. Previous work has been done on studying the full dynamics of the system and the control strategies that drive the vehicle to the desired planned trajectories [41].

*Assumption 2*: the propulsion system of the HAUV maintains a constant thrust at economic power consumption, i.e., the vehicle maintains a constant flight speed $V_{air}$ in the air, and a constant operation speed $V_{sea}$ during the underwater navigation phase.

*Assumptions 3*: the Information Map (IM) of environmental features studied in this paper is given based on the actual observation needs of marine scientists, so the planning problem in this paper is based on the prior known IM.

The information-driven path planning problem studied in this paper is formulated as follows: To study the characteristics of the sea-air interface in a specific sea area, a HAUV carrying limited energy $E_{max}$ is deployed from the deck of a research vessel. It is commanded to perform a sampling mission and at a specified mission time $T_{max}$ must return or land to another base station. According to Assumption 2, the velocity of the HAUV while flying in the air is $V_{air}$, the velocity while navigating underwater is $V_{sea}$, and the global sampling path is $\mathbb{P} = \{\mathcal{P}_1, \mathcal{P}_2, \ldots, \mathcal{P}_h\}$, where $h$ is the number of discrete path points. In summary, the mathematical model can be established in the following form:

$$\mathbb{P}^* = argmax\, f_\tau(IM, V_{air}, V_{sea}, V_c, C_{obs}, E_{max}, T_{max}, \mathbb{W})$$

$$s.t. \quad V_{air} = 0, V_{sea} = 0, \quad\quad\quad (1)$$
$$\forall i \in \{1,2,\ldots,h\}, \ \mathcal{P}_i \notin C_{obs}$$
$$E \leq E_{max},\ T \leq T_{max},$$

Where $f_\tau()$ is the information collection function that returns the total amount of information collected along the entire path. $\mathbb{W}$ is the 3D workspace of the HAUV. $E$ is the total energy consumption of the HAUV along the path $\mathbb{P}$ when performing the mission. $T$ is the total time of the HAUV along the path $\mathbb{P}$.

### 3.1 Optimization criterion

The optimization criterion of the HAUV information-driven path planning problem is that the optimized path can maximize the information collected within a specific mission area and limited budgets. The information collection relies on the sensors onboard the HAUV, which can detect and collect data within a certain range of the current location $\mathcal{P}_i$, and build a 3D array $measured[]$ with all initial values of zero, the same size as the IM, which holds the values of collected environmental feature information. The HAUV performs the sampling task along the optimized path, constantly updating the information storage array $measured[]$. In addition, the value of feature information of different spatial locations in the air and sea environment may be variable, introducing the weight coefficient $\kappa$ into the information acquisition function. In summary, the total amount of information collected throughout the path can be expressed in the following form:

$$f_\tau(P) = \sum_{j=1}^{J} \kappa_j \cdot measured[\rho_j] \quad\quad\quad (2)$$

$$s.t. \quad \rho_j \in W, j \in \{1,2,3,\ldots,J\}$$

In which, $\rho_j$ is the coordinate position of a raster point in the HAUV workspace. $\kappa_j$ is the information value weight of raster point $\rho_j$. $J$ is the number of discrete raster points in the 3D workspace.

### 3.2 Sensor models

HAUVs can carry a variety of sensors for aerial, surface, and underwater phenomena observation and information acquisition. Different sensors may have different sensing ranges and information acquisition capabilities, recent research makes use of continuous measurements of local ocean conditions from on-board current profiling sensors mounted in HAUV, e.g., an Oculus M750D forward looking sonar [42] and a StarFish sidescan sonar for information collection [43]. The Oculus M750D sensor is composed of 512 beams that allow aperture up to 120 meters in front of the HAUV. In this paper, the complexity of the sensor model is simplified. According to previous studies on various types of sensors, the ability of sensors to collect feature information decays with increasing distance [44, 45]. Therefore, the HAUV's ability to collect information about the surrounding workspace at point $\rho_j$ can be expressed in the following form:

$$\mathcal{A}(\mathcal{P}_i, \rho_j) = \begin{cases} \mathcal{A}_{d_{max}} e^{-\sigma(\frac{d_j}{d_{max}})^2} , \text{if } d_j \leq d_{max} \\ 0 \quad , \text{if } d_j > d_{max} \end{cases} \quad (3)$$

$$\text{s.t.} \quad \mathcal{A}_{d_{max}} \in [0,1]$$

In which, $d_j$ is the Euclidean distance between the two points $\mathcal{P}_i$ and $\rho_j$; $\mathcal{A}_{d_{max}}$, $\sigma$ and $d_{max}$ are the parameters of the sensor model, which control the sensing range and the sensor capability. Then, the sensor at the path point $\mathcal{P}_i$ can collect the amount of information at the point $\rho_j$ in the workspace as:

$$\text{sensor}(\mathcal{P}_i, \rho_j) = \text{IM}(\rho_j) \cdot \mathcal{A}(\mathcal{P}_i, \rho_j) \quad (4)$$

The HAUV navigates along a planned path, the information about environmental features is continuously collected by the sensors, which can continuously update the information storage array $measured[]$ established in subsection 3.1. If the array $measured[\rho_j]$ stores the information value of the collected raster points $\rho_j$ is less than the information value collected by the sensor $\text{sensor}(\mathcal{P}_i, \rho_j)$, the array $measured[\rho_j]$ is updated. Otherwise, it is not updated.

$$measured[\rho_j] = \begin{cases} \text{sensor}(\mathcal{P}_i, \rho_j) , \text{if measured}[\rho_j] \leq \text{sensor}(\mathcal{P}_i, \rho_j) \\ measured[\rho_j] , \text{else} \end{cases} \quad (5)$$

The update detection is carried out during the sampling task until the end of the path. Finally, the information collection function Equation (1) returns the total amount of information that can be collected for the whole path.

### 3.3 Constraint conditions

Constraints on the HAUVs include the fact that the vehicles carry a limited amount of energy per mission and the possibility of pre-specified mission times by marine scientists.

### 3.3.1 Energy constraints

The HAUV has three different modes of motion during the sampling mission: airborne mode, underwater navigation mode, and cross-media mode. A complete mathematical relationship between speed and energy consumption of the HAUV in these three modes of motion is not yet available. According to the existing literature, the relationship between speed and energy consumption of unmanned vehicles (e.g., UAVs and AUVs) can be analyzed. When the economic speed is maintained, the vehicle's energy consumption per unit time can be minimized. Combining with assumption 3, this paper simplifies the mathematical model of the speed-energy consumption relationship of the HAUV, in which the power of the HAUV for economical air flight is known as $P_{air}$ And the power for economic underwater navigation is $P_{sea}$. The total power consumption of the HAUV can be expressed as the sum of the power consumption in the flight mode $E_{air}$, the power consumption in the underwater navigation mode $E_{sea}$ and the power consumption in the cross-media transition $E_{switch}$.

$$E = E_{air} + E_{sea} + E_{switch} \quad (6)$$

$$E_{air} = P_{air} \cdot T_{air} \quad (7)$$

$$E_{sea} = P_{sea} \cdot T_{sea} \quad (8)$$

$$\text{s.t.} \quad E \leq E_{max}$$

In addition, the total energy consumed by the HAUV per mission $E$ cannot exceed the maximum energy $E_{max}$. In equations (7), (8) $T_{air}$ and $T_{sea}$ are the operating time of the HAUV in air flight mode and underwater navigation mode. The solution about the operating time will be explored in subsection 3.3.2. It should be noted that the time and power consumption of a single cross-media transition is simplified to a constant.

3.3.2 Mission time Constraints

Typically, marine scientists expect HAUV to complete sampling missions within a specified mission time. The total mission time can be expressed in the following form.

$$T = T_{air} + T_{sea} + T_{switch} = \sum_{i=1}^{h-1} \frac{|\mathcal{P}_i - \mathcal{P}_{i+1}|}{V_{abs\_i}} \tag{9}$$

$$\text{s.t.} \quad T \leq T_{max}, \quad i \in \{1, 2, \dots, h-1\}$$

In which, $V_{abs\_i}$ is the actual operational velocity of the HAUV in the inertial coordinate system. The velocity of the HAUV is $V_{hauv}$, including the $V_{air}$ in airborne flight mode, the $V_{sea}$ in the underwater navigation mode and the variable speed motion in the cross-media transition mode. Considering the air-sea environment, there are wind and flow fields, the actual operating velocity $V_{abs}$ of the HAUV in the inertial coordinate system is affected by the environment. The actual operational velocity of the vehicle will be solved by the velocity vector synthesis method [46].

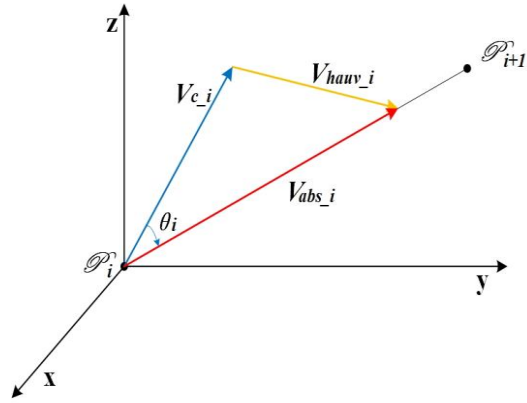

Figure 2 Schematic diagram of velocity synthesis

As shown in Figure 2, the actual operational velocity direction of the HAUV on the path segment $\mathcal{P}_i\mathcal{P}_{i+1}$ should be consistent with the forward direction and the angle $\cos\theta_i$ satisfies the following equation.

$$\cos\theta_i = \frac{V_{c\_i} \cdot V_{abs\_i}}{\|V_{c\_i}\|\|V_{abs\_i}\|} = \frac{V_{c\_i} \cdot a_i}{\|V_{c\_i}\|} = \frac{u_{c\_i}a_{x\_i} + v_{c\_i}a_{y\_i} + w_{c\_i}a_{z\_i}}{\sqrt{u_{c\_i}^2 + v_{c\_i}^2 + w_{c\_i}^2}} \tag{10}$$

In which, $a_i$ is the unit vector of $V_{abs\_i}$, that is, the unit vector of the path segment $\mathcal{P}_i\mathcal{P}_{i+1}$ whose components in the x, y, and z components in the three directions are $a_{x\_i}, a_{y\_i}, a_{z\_i}$.

Suppose that the path planning system has in advance the complete distribution information of the velocity field, including the velocity direction and magnitude. Then, the direction and magnitude of the velocity field $V_{c\_i}$, the direction of the actual operating velocity $V_{abs\_i}$ of the HAUV in the inertial coordinate system, and the velocity generated by the thrusters in the airframe coordinate system $V_{hauv\_i}$, according to the Cosine theorem for triangles:

$$V_{c\_i}^2 + V_{abs\_i}^2 - 2V_{c\_i}V_{abs\_i}\cos\theta_i = V_{hauv\_i}^2 \tag{11}$$

The actual magnitude of the operational speed of the HAUV in the inertial coordinate system $V_{abs\_i}$ can be deduced as the quadratic solution of equation (12).

$$V_{abs\_i}^2 - 2(u_{c\_i}a_{x\_i} + v_{c\_i}a_{y\_i} + w_{c\_i}a_{z\_i})V_{abs_i} + V_{c\_j}^2 - V_{hauv\_i}^2 = 0 \tag{12}$$

Let $\triangle = 4(u_{c\_i}a_{x\_i} + v_{c\_i}a_{y\_i} + w_{c\_i}a_{z\_i})^2 + 4V_{hauv\_i}^2 - 4V_{c\_j}^2$, when $\triangle<0$, this equation has no real number solution; when $\triangle \geq 0$, the solution of this equation can be expressed as.

$$V_{abs\_i} = u_{c\_i}a_{x\_i} + v_{c\_i}a_{y\_i} + w_{c\_i}a_{z\_i} \pm \frac{1}{2}\sqrt{\triangle} \tag{13}$$

When there are two feasible solutions, the positive solution with the larger value is usually chosen as the value of $V_{abs\_i}$. In addition, when the value of the resulting solution is zero or negative, i.e., the value of $V_{abs\_i}$ is not positive. This indicates that the path segment $\mathcal{P}_i\mathcal{P}_{i+1}$ is not reachable. A new feasible path needs to be generated. When $\triangle < 0$, it means the velocity component of the HAUV in the path segment $\mathcal{P}_i\mathcal{P}_{i+1}$ direction is not sufficient to counteract the velocity component of the flow or wind field in this direction, resulting in no real number solution to equation (12). Therefore, the algorithm must solve for the actual operating speed of the HAUV in each path segment, check whether the HAUV can reach each path point, and ensure that the optimized sampling paths are feasible.

3. 4 Path formation and smoothing

The information-driven path planning algorithm usually outputs a set of discrete path nodes $\{p_1, p_2, p_3, ...\}$. To generate a path that satisfies the kinematics and dynamics of the HAUV, this paper adopted B-spline curves for path smoothing [47, 48]. The principle of the B-spline curve is as follows. Assuming that the information-driven path planning algorithm generates six path nodes after optimization points $\{p_1, p_2, p_3, p_4, p_5, p_6\}$, where $p_1$ is the starting point and $p_6$ is the end point. These six path nodes are used as control points for the B-spline curve for curve fitting.

$$P(s_k) = \sum_{n=0}^{N} p_{k+n} B_{n,N}(s_k) \tag{14}$$

$$\text{s.t.} \quad s_k \in [0,1], k \in [1,2,...,6]$$

Where $N$ is the order of the B-spline curve, $B_{n,N}(s_k)$ is the Bernstein fundamental polynomial representing the B-spline basis function of the curve, which is defined as follows.

$$B_{n,N}(s_k) = C_N^n s_k^n (1-s_k)^{N-n} = \frac{N!}{n!(N-n)!} s_k^n (1-s_k)^{N-n}, n \in \{0,1,...,N\} \tag{15}$$

When $N = 3$, $P(s_k)$ is continuously second-order derivable, i.e., the cubic B-spline curve generates a smooth path with continuous velocity and acceleration variation patterns from the start to the end of the HAUV. Therefore, the output optimized path $\mathbb{P} = \{\mathcal{P}_1, \mathcal{P}_2, ..., \mathcal{P}_h\}$ is continuous, smooth, and feasible.

3.5 Sea and air 3D environment modelling

The ocean and atmosphere environmental models can be obtained from official forecasts or built based on analytical equations.

3.5.1 Forecast-based 3D environment model for air and sea

At present, low-resolution hydrometeorological parameter data and environmental velocity field data can be obtained through ocean observation networks, satellite measurements, etc. The hydrometeorological parameter data of interest are used to construct information maps of sea-air environment characteristics and input into the path planning system of HAUVs together with wind and currents field data so that HAUV can perform the information sampling task to obtain higher resolution and finer hydrometeorological parameters in the target sea area.

The National Oceanic and Atmospheric Administration (NOAA) website provides forecast data for all types of hydrometeorological parameters. The Regional Navy Coastal Ocean Model (NCOM) datasets provide forecasts of ocean temperature, salinity, and horizontal currents at different depths with a horizontal resolution of about 3 km.

This paper downloads hydrometeorological parameters and targets from the website mentioned above. Wind and current fields in the sea area are analyzed, and information such as ocean salinity and atmospheric humidity is extracted. The data are fused to establish the forecast-based sea-air environment characteristic information map IM and the model of the sea-air environment velocity field $V_c$. Two points need to be clarified here. First, both ocean and atmospheric models only provide data on flow and wind velocities in the horizontal direction and combined with studies in the existing literature, the velocity components $w_c$ of the wind and flow fields in the vertical direction are very small compared to the velocity components in the horizontal direction $u_{c\_i}$ and $v_c$ are very small in the order of magnitude and therefore negligible[49]. Second, the forecast-based air-sea models are mainly for studies of large sea areas at the several-kilometer level. In contrast, interpolation is required to generate higher resolution maps for studies of sea areas at the 100-meter level or smaller scales. In this case, instead, a mathematical type of the 3D environment of air and sea in small-scale sea areas is established based on analytical equations.

3.5.2 Analytic equation-based 3D environment model for air and sea

The distribution of 3D environmental feature information usually conforms to a ternary Gaussian distribution. Assuming that there are *B* feature information regions of interest to marine scientists in the target area, the synthesized 3D *IM* can be represented as a Gaussian mixture model.

$$IM = \sum_{b=1}^{B} g_b \cdot \mathcal{N}(\mu_b, \Sigma_b) \tag{16}$$

$$\text{s.t.} \quad b \in \{1, 2, 3, \ldots, B\}$$

Where $\mu_b = [x_b, y_b, z_b]$ is the mean value, which represents the position of the center of the feature information b in the workspace; $\Sigma_b$ is a covariance matrix of size 3×3, which controls the dispersion of the feature information b in the x, y, and z directions; $g_b$ is the weight parameter, which controls the peak size of the feature information b.

In the simulation experiments of this paper, the maps of the environmental feature information based on the analytic equations are created randomly by the Gaussian mixture model. The Gaussian distribution of feature information b in 3D space is used as an example. A point in the workspace is randomly selected as the center $\mu_b$ of feature information b, and the randomly generated covariance matrix $\Sigma_b$ is constructed in the following way.

(1) First construct a random diagonal matrix $\mathcal{A} = \text{diag}([\mathcal{A}_x, \mathcal{A}_y, \mathcal{A}_z])$, where, $\mathcal{A}_x, \mathcal{A}_y, \mathcal{A}_z$ is random positive;

(2) Rebate a random matrix $\mathcal{B} = \text{Rand}(3,3)$, figure out the standard orthogonal group of matrix $\mathcal{B} = \text{orth}(\mathcal{B})$;

(3) If the characteristic value of the matrix $\mathcal{C} = \mathcal{B}^T \mathcal{A} \mathcal{B}$ is greater than or equal to 0, the matrix is a randomly generated symmetric semi-positive matrix, which can be used as a covariance matrix $\Sigma_b$.

There is B randomly generated environmental characteristic information in the workspace. Then, the information value of any point $\rho_j$ can be defined as:

$$I(\rho_j) = \sum_{b=1}^{B} \frac{g_b}{\sqrt{(2\pi)^3 |\Sigma_b|}} e^{-\frac{1}{2}(\rho_j - \mu_b)^T \Sigma_b^{-1}(\rho_j - \mu_b)} \tag{17}$$

To have a unified measurement standard for different target sea areas, we used normalized processing between the $I_{air}$ and $I_{sea}$ in this paper. It means the data distribution range of the feature information $I_{air}$ and $I_{sea}$ of all grid points in the workspace is [0, 1]. According to the result discussed in subsection 3.5.1, the 3D velocity field can be decomposed into a set of different heights and depths of the 2D horizontal speed field associated with each other. The speed field model in the 2D level can be established and superimposed in a plurality of viscous Lamb eddy [50]. According to the analysis equation of the Lamb vortex, a Lamb vortex in the horizontal direction of the vertical position can be expressed as follows.

$$V_{c\_xy} = f_c(\mathbb{R}_i^o, \eta, \zeta) \tag{18}$$

$$u_c(\mathbb{R}_i) = -\eta \frac{y - y_o}{2\pi (\mathbb{R}_i - \mathbb{R}_i^o)^2} \left[1 - e^{-\left(\frac{(\mathbb{R}_i - \mathbb{R}_i^o)^2}{\zeta^2}\right)}\right] \tag{19}$$

$$v_c(\mathbb{R}_i) = \eta \frac{x - x_o}{2\pi (\mathbb{R}_i - \mathbb{R}_i^o)^2} \left[1 - e^{-\left(\frac{(\mathbb{R}_i - \mathbb{R}_i^o)^2}{\zeta^2}\right)}\right] \tag{20}$$

Among them, $\mathbb{R}_i = \begin{bmatrix} x \\ y \end{bmatrix}$ indicates a 2D working space, $\mathbb{R}_i^o = \begin{bmatrix} x_o \\ y_o \end{bmatrix}$ indicates the center position of the vortex, η represents vortex strength, ζ and indicates the radius of the vortex. If there are multiple different locations, intensity and radius of Lamb vortex in the field, the above three formulas are superimposed to solve the size and direction of the horizontal direction velocity field $V_{c\_xy}$. The 3D ambient speed field continuously gradient in the vertical direction is created by introducing the argument $\mathbb{R}_i^o, \eta, \zeta$.

Figure 3 shows the maps of sea-air environment feature information generated based on the Gaussian mixture model and the velocity field generated by Lamb vortex. The size of the raster in schematic diagram is 100×100×13, indicates the searching area of 5km×5km×600m, and the gradient transparent red color indicates the information value of atmospheric features from 1 to 0, the gradient transparent blue color indicates the information value of ocean features from 1 to 0. The translucent dark blue plane indicates the sea level. In this paper, the wind field velocity is controlled within 5m/s, and the flow field velocity is controlled within 0.4m/s. From the top view in Fig.3b, we can see the velocity field in the vertical direction.

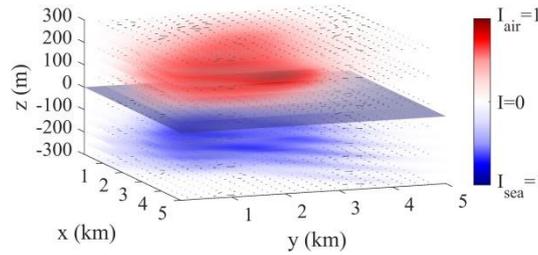

a) Information on the characteristics of the sea and air 3D environment and the distribution of velocity fields

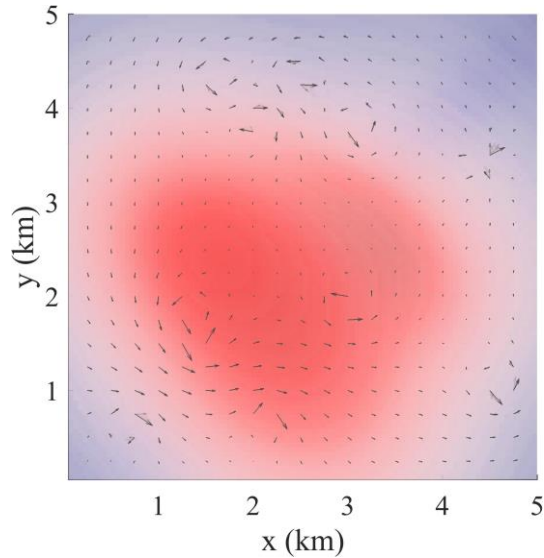

b) Top view

Figure 3 Schematic diagram of the 3D marine-atmospheric environment model

4. INFORMATION-DRIVEN PATH PLANNING ALGORITHM DESIGN

The objective of the RAST* algorithm is to optimize the path of the HAUV, which maximizes the collection of air-sea environmental features and autonomously allocate the tasks of the HAUV in the air and underwater to meet the constraints of limited energy and preset mission time of the vehicle, and to avoid obstacles effectively. The RAST* algorithm innovatively combines the sampling strategy based on the tournament point selection method, the information heuristic search process, and the RRT* algorithm framework to achieve an efficient search of the air-sea information map to solve the optimal sampling path quickly. Four versions of the RAST* algorithm are designed to verify the effectiveness and superiority of the RAST* algorithm and study the effects of different optimization methods on the computational speed and solution accuracy of the RAST* algorithm. The deformed Rapidly-exploring Random Sampling Tree* (RRST*) algorithm is designed according to the different sampling strategies. The RAST*-I/E algorithm and the RAST*-I algorithm are designed according to the information heuristic search process; the deformed Rapidly-exploring Adaptive Sampling Tree (RAST) algorithm is designed according to the presence or absence of information heuristic search and the reshaping process of parent nodes. Meanwhile, this paper also uses the classical fast search information gathering tree algorithm and the PSO algorithm as comparison algorithms. The optimization process and technical details of these six algorithms are discussed in detail in the following.

4.1 Rapidly-exploring Adaptive Sampling Tree Algorithm

The RAST* algorithm is a sampling-based algorithm inspired by the RRT* algorithm, but what differs from the RRT* algorithm is the introduction of a sampling strategy based on the tournament point selection method and an information heuristic search process in the main structure of the algorithm. Based on the RAST* algorithm, the sampling strategy tends to grow branches to the regions with high feature information values. The information heuristic search process searches for global sampling paths with low energy consumption during the iterative process of the algorithm, which helps to avoid the RAST* algorithm from falling into local optimum solutions. These two improvements enable the RAST* algorithm to generate the global optimal path for the HAUV. The pseudo-code of the RAST* algorithm is shown in Algorithm 1, and the main flow is as follows.

First, the parameters that need to be inputted before the RAST* algorithm can be executed include.

- Environment model parameters $PA_e$: information map IM, wind/currents field $V_c$, obstacle $C_{obs}$;
- HAUV related parameters $PA_{hauv}$: the aerial speed $V_{air}$ and the operation power $P_{air}$, underwater speed $V_{sea}$ and the operation power $P_{sea}$, cross-media energy consumption $E_{switch\_1}$ and time $T_{switch\_1}$, limited energy $E_{max}$, sensor parameters $\mathcal{A}_{d_{max}}, \sigma, d_{max}$;
- Task-related parameters $PA_m$: Task start location $q_{init}$, end location $q_{final}$, Preset mission time $T_{max}$;
- RAST* algorithm parameters $PA_{cdrast}$: the number of tourists M, step size δ, neighboring radius r, maximum iterative number Max_it, the number of iterations the result is no longer improved It_stop.

Let Tree = (Vertex, Edge) denotes the adaptive sampling tree, Vertex is the set of tree nodes, and Edge is the set of tree branch segments formed by node connections. Best_IG is a variable storing the optimal amount of information, Bestsol is a one-dimensional array storing the optimal amount of information for each iteration.

**Algorithm 1** Rapidly-exploring Adaptive Sampling Tree*

**Enter:** 3D environment model parameters $PA_e$, HAUV related parameters $PA_{hauv}$, task-related parameters $PA_m$, CDRAST* algorithm parameters $PA_{rast}$.

1: Vertex ← $\{q_{init_m}\}$; Edge ← ∅; Tree = (Vertex, Edge); Best_IG = 0; Bestsol(1) = 0;
2: **for** it = 1 to Max_it **do**
3:    $q_{ts}$ ←TournamentSample(IM, $\mathbb{M}$);
4:    $q_{nearest}$ ←Nearest($q_{ts}$, Vertex);
5:    $q_{new}$ ←Steer($q_{nearest}, q_{ts}, r$);
6:    **if** CollisionFree($C_{obs}, q_{nearest}, q_{new}$) **then**
7:    $Q_m$ ←Near(Vertex, $q_{new}, r$);
8:    $c_{max} = 0$;
9:    **for** each $q_m \in Q_m$ **do**
10:    [IG, E, T, $\mathbb{P}$] = FitnessFun($q_{new}, q_m, q_{init}, q_{final}$, Vertex, $PA_{hauv}, PA_e$);
11:    $c1 = \frac{IG}{E}$;
12:    **if** $c1 \geq c_{max}$ & E≤ $E_{max}$ & T ≤ $T_{max}$ & CollisionFree($C_{obs}, \mathbb{P}$) **then**
13:    $c_{max} = c1$; $q_{max} \leftarrow q_m$; $\mathbb{P}_{max} \leftarrow \mathbb{P}$;
14:    **end if**
15: **end for**
16: **if** CollisionFree($C_{obs}, \mathbb{P}$) **then**
17: $q_{new}$.Parent ← $q_{max}$;
18: $q_{new}$.T ← $f_{Time}(\mathbb{P}_{max})$;
19: $q_{new}$.E ← $f_{Energy}(\mathbb{P}_{max})$;

20: $q_{new}.IG \leftarrow f_{Energy}(\mathbb{P}_{max})$;

21: Vertex $\leftarrow$ Vertex $\cup \{q_{new}\}$

22: Edge $\leftarrow$ Edge $\cup \{(q_{max}, q_{new})\}$

23: **if** $q_{new}.IG >$ Best_IG **then**

24: Best_IG $= q_{new}.IG$

25: **end if**

26: **end if**

27: **end if**

28: Bestsol(it) $=$ Best_IG

29: **if** Bestsol(it) $-$ Bestsol(it) $=$ Best_IG

30: break:

31: **end if**

32: **end for**

33: **return** Tree $=$ (Vertex, Edge);

**Output**: $\mathbb{P}^*$, **M_Best_IG**, Bestsol

---

**Algorithm 2** Tournament point selection function

1: **function** TOURNAMENTSAMPLE(*IM*, $\mathbb{M}$)

2:    randomly selected $\mathbb{M}$ raster points $\{\varrho_1, \varrho_2, ..., \varrho_\mathbb{M}\}$ from IM;

3:    $q_{ts} \leftarrow \varrho_1$

4:    for i=2 to $\mathbb{M}$ do

5:      if $IM(\varrho_j) > IM(q_{ts})$ **then**

6:        $q_{ts} \leftarrow \varrho_j$;

7:      **end if**

8:    **end for**

9:    **return** $q_{ts}$;

10: **end function**

---

**Algorithm 3** Nearest point selection function

1: **function** NEAREST ($q_{ts}$, Vertex)

2:    vn=Length(Vertex);

3:    for i =1 to vn do

4:      Dis(i) = Distance($q_i$, $q_{ts}$);

5:    **end for**

6:    $[Dis_{min}, Index_{min}] = \min(Dis)$;

7:    **return** $q_{Index_{min}}$

8: **end function**

**Algorithm 4** Steering function

1: **function** STEER($q_{nearest}$, $q_{ts}$, $\delta$)
2:     D=Distance($q_{nearest}$, $q_{ts}$);
3:     **if** D>$\delta$ **then**
4:         $q_{new} = q_{nearest} + (q_{ts} - q_{nearest}) * \delta/D$
5:     **else**
6:         $q_{new} \leftarrow q_{ts}$
7:     **end if**
8:     **return** $q_{new}$
9: **end function**

The main structure of the RAST* algorithm has four key processes as follows, and the whole process is shown in Figure 4.

**Sampling strategy based on tournament point selection method** (Algorithm 1, line 3): the tournament point selection method replaces the original random sampling method in the RRT* algorithm to generate sampling points. The TOURNAMENTSAMPLE() function in the steps shown in Algorithm 2 and Figure 4a. Raster points $\{\varrho_1, \varrho_2, ..., \varrho_\mathbb{M}\}$ are randomly selected from IM. Compare the feature information values corresponding to these $\mathbb{M}$ raster points and return the raster point with the largest value as the sampling point $q_{ts}$.

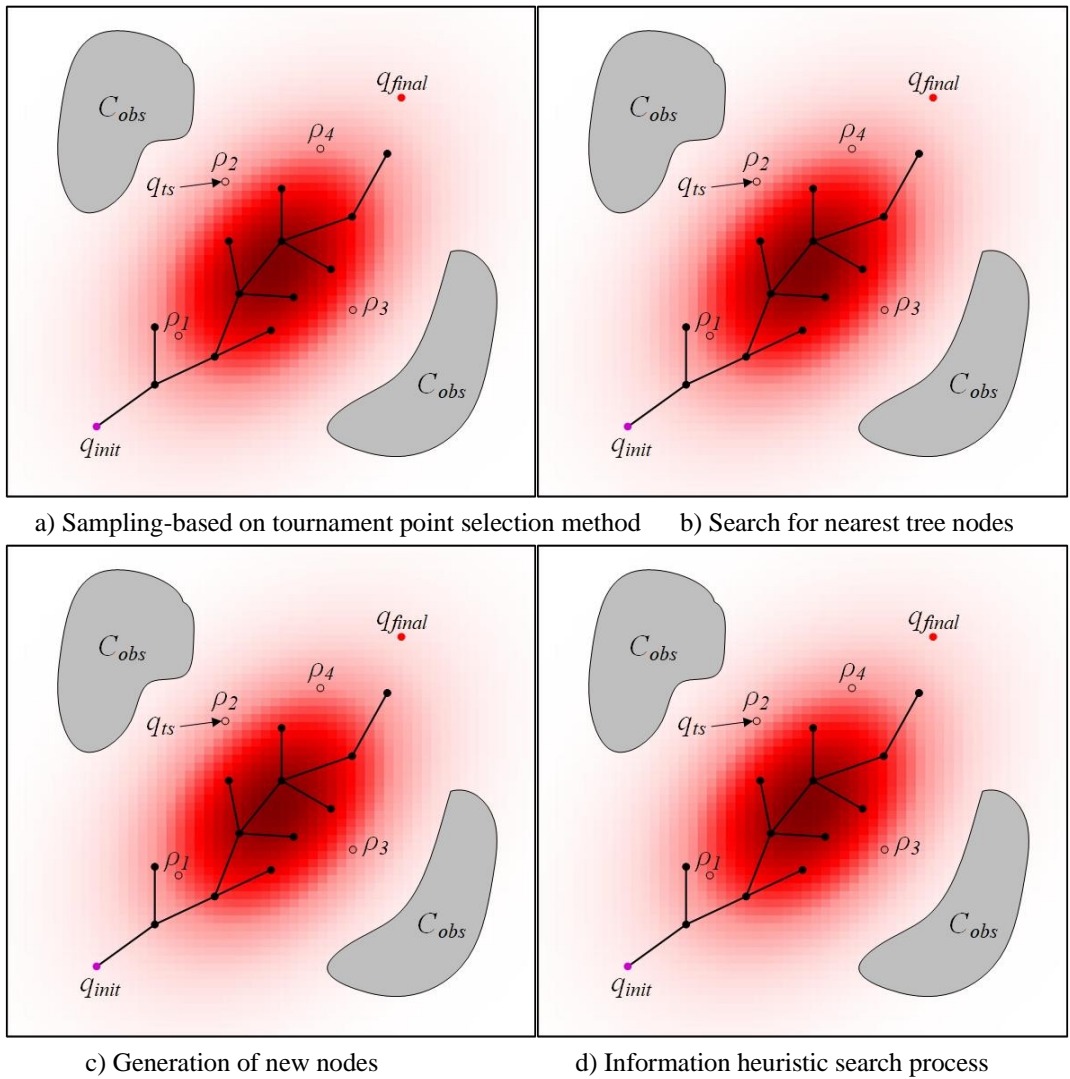

a) Sampling-based on tournament point selection method     b) Search for nearest tree nodes

c) Generation of new nodes     d) Information heuristic search process

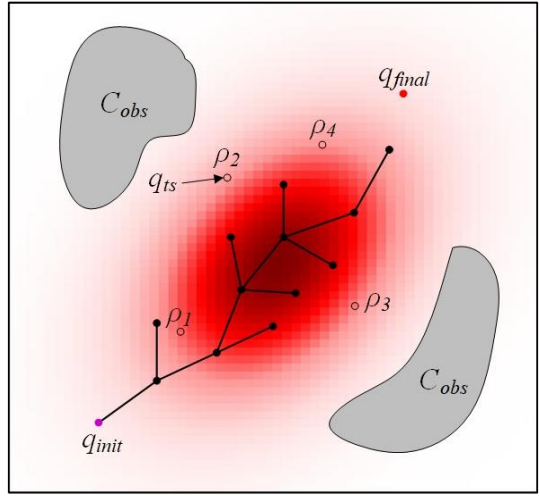

e) Parent node reshaping

Figure 4 An illustration of RAST*

---

**Algorithm 5** Collision detection function

---

1: **function** COLLISIONFREE($C_{obs}$,varargin)
2:   **if** varargin is not part of the $C_{obs}$ **then**
3:     **return** 1
4:   **else**
5:     **return 0**
6:   **end if**
7: **end function**

---

**Algorithm 6** Find the set of nearby nodes function

---

1: **function** NEAR (Vertex, $q_{new}$, r)
2:   $\mathbf{Q_m} \leftarrow \emptyset$
3:   vn=Length(Vertex)
4:   **for** i = 1 to vn **do**
5:     **if** Distance($q_i$, $q_{new}$)< r **then**
6:       $\mathbf{Q_m} \leftarrow \mathbf{Q_m} \cup q_i$
7:     **end if**
8:   **end for**
9:   **return** $\mathbf{Q_m}$
10: **end function**

---

**Algorithm 7** Adaptation function

---

1: **function** FITNESSFUN($q_{new}$, $q_m$, $q_{init}$, $q_{final}$, Vertex, $PA_{hauv}$, $PA_e$)
2:   $\mathbb{P} = \text{Connection}(q_{new}, q_m, q_{init}, q_{final}, \text{Vertex})$;
3:   $IG = f_\tau(\mathbb{P})$;
4:   $E = f_{Energy}(\mathbb{P})$;
5:   $T = f_{Time}(\mathbb{P})$;
6:   **return** $[IG, E, T, \mathbb{P}]$
7: **end function**

**Search for the nearest tree node and generation of new nodes** (Algorithm 1, lines 4-5): the nearest tree node is found in the set of tree nodes Vertex around the sampling point $q_{ts}$. The search process of the nearest tree node is represented as the Nearest() function, and the specific steps are shown in Algorithm 3 and Figure 4b. Length(Vertex) function returns the number of tree nodes vn in Vertex. The Distance() function is used to solve for the Euclidean distance between two tree nodes, finding the index of the nearest tree node in the set Vertex, based on the one-dimensional array Dis. Use the tree node returned by the Nearest() function as $q_{nearest}$. Nearest() function can also be expressed in the following:

$$\text{Nearest}(q_{ts}, \text{Vertex}) = \underset{q_i \in \text{Vertex}}{\arg\min} |q_{ts} - q_i|, \quad i=1,2,\ldots,vn \tag{21}$$

Then, according to the steering function Steer(), grows from the tree node $q_{nearest}$ to the sampling point $q_{ts}$ with the tree branch whose length is $\delta$ and generate a new node $q_{new}$, as shown in Figure 4c. If the Euclidean distance between the two points $\{q_{nearest}, q_{ts}\}$ is shorter than a step $\delta$. Then $q_{ts}$ is the new node $q_{new}$. Steer() function can also be expressed in the following:

$$\text{Steer}(q_{nearest}, q_{ts}, \delta) = \begin{cases} q_{nearest} + \dfrac{\delta}{|q_{ts} - q_{nearest}|} \cdot (q_{ts} - q_{nearest}), & \delta \le |q_{ts} - q_{nearest}| \\ q_{ts}, & \delta \ge |q_{ts} - q_{nearest}| \end{cases} \tag{22}$$

**Information heuristic search process** (Algorithm 1, lines 7-15): after checking that nodes $q_{nearest}, q_{new}$ are not in the obstacle space. find the tree node whose distance between node $q_{new}$ and itself is less than r in the set of tree nodes Vertex according to the Near() function and store it in the set of neighbors $Q_m$. The Near() function can also be expressed in the following:

$$\text{Nearest}(q_{ts}, \text{Vertex}, r) = q_i \in \text{Vertex}: |q_{ts} - q_i|, \quad i=1,2,\ldots,vn \tag{23}$$

Using Connection() for each node in the neighborhood set $Q_m$. First retraces the parent node of $q_m$ until it returns to the starting point $q_{init}$ and then use a B-spline curve to fit the path segmentation from the starting point $q_m$ through the tree nodes to $q_{new}$ and finally to the $q_{final}$, until it forms the curvature continuous sampling path $\mathbb{P}$. According to Equations (1), (6) and (9), the total amount of collected information, the total energy consumption and the total time consumption of path $\mathbb{P}$ can be solved. By cycling, the path that satisfies the constraints of energy, mission time and no collision along with the most amount of information collected per unit power consumption is found as $\mathbb{P}_{max}$, and the neighboring nodes that constitute this path are recorded $q_{max}$, as shown in Figure 4d.

**Remodelling of the parent node and update of the optimal solution** (Algorithm 1, lines 16-31): after checking that all discrete points in the path $\mathbb{P}_{max}$ are not within the obstacle space, the $q_{max}$ will be recorded as the parent of $q_{new}$, storing the total amount of collected information, total energy consumption and total time consumption of the path $\mathbb{P}$ constructed by node $q_{new}$, and add nodes $q_{new}$ to the set of tree nodes Vertex, segment ($q_{max}, q_{new}$) will be added in the tree branch set Edge as shown in Figure 4e. If the amount of collected information by the path generated in this iteration is greater than that of the previous iteration, then update the variable Best_IG as the optimal solution for this iteration. Otherwise do not update. After that, Best_IG is added to the array Bestsol. The algorithm keeps iterating in a loop until the optimal solution has no improvement after It_stop iterations. Otherwise, it iterates until the maximum number of iterations Max_it and outputting the path $\mathbb{p}^*$, the optimal solution Best_IG and the array of optimal solutions Bestsol generated by each iteration.

It is important to note here that the path planning problem studied in this paper has a large number of constraints. According to the characteristics of the HAUV model, if the mission start and end points are particularly far apart, there may be a situation where the HAUV is unreachable under the constraints of limited energy and mission time. In this case, the RAST* algorithm method may not be able to generate a feasible solution after It_stop iterations, and output Best_IG=0. The optimized path $\mathbb{p}^*$ is empty. In this case, the mission start and end positions need to be set again reasonably.

4.1.1 Different improved forms of the RAST* algorithm

The RAST* algorithm uses $c_{max}$ as the heuristic factor (line 11 of Algorithm 1). $c_{max}$ is the amount of information collected by the path per unit power consumption to judge which node in the neighborhood set $Q_m$ will be selected as the parent node of $q_{new}$. Hence, this algorithm is named as RAST*-I/E algorithm in this paper. The corresponding counterpart is the RAST*-I algorithm, it heuristic factor $c_{max}$ is the total amount of information collected throughout the path to judge which node in the neighborhood set $Q_m$ will be selected as be the parent node of $q_{new}$. That is, the RAST*-I algorithm directly uses the target value as the value of the $c_{max}$. In theory, if the total amount of collected information is set as a heuristic factor and the power consumption as a constraint throughout the path, the initial search process may focus on acquiring more information in the short term without limiting power consumption. As a result, in the subsequent iterations, the tree nodes will not be able to grow to the region with high feature information value due to the lack of energy. Thus the RAST*-I algorithm tends to fall into local optimal solutions. The subsequent simulation experiments will compare the RAST*-I/E algorithm with the RAST*-I algorithm to analyze the information heuristic search process and analyze the influence of the selection of the heuristic factor $c_{max}$ on the final results.

The framework of the RAST algorithm is consistent with that of the RRT algorithm. There is no heuristic search and parent node reshaping process in the optimization process and the tree node $q_{nearest}$ is directly used as the parent node of $q_{new}$, it means that lines 7-20 of Algorithm 1 are not executed. Although the RAST algorithm introduces a sampling strategy based on the tournament point selection method, leading the adaptive sampling tree to grow to regions with higher values, the RAST algorithm lacks inspiration. It does not re-evaluate the fitness of the tree nodes with the newly generated nodes, so only feasible solutions can be obtained after optimization. The optimality of the RAST algorithm is the same as that of the RRT algorithm. Subsequent simulation experiments will compare the RAST*-I/E algorithm with the RAST algorithm to analyze the necessity and importance of applying the information heuristic search and the reshaping process of the parent nodes.

In addition, the RRST* algorithm is designed in this paper depending on the sampling strategy. The sampling strategy of the RRST* algorithm is the same random sampling strategy as the RRT* algorithm, i.e., a point is randomly selected as the sampling point $q_{ts}$ (line 3 of Algorithm 1), i.e.𝕄=1; the rest of the RRST* algorithm procedure is the same as the RAST*-I/E algorithm. Theoretically, the advantage of the RRST* algorithm is that it maintains the randomness of the algorithm and can search and grow branches randomly. Hence, the RRST* algorithm explores the whole space more comprehensively. However, due to the stochastic nature of the RRST* algorithm, it requires more iterations to converge, leading to a longer computation time than the RAST*-I/E algorithm. Subsequent simulations experiments will also compare the RAST*-I/E algorithm and the RRST* algorithm to analyze the effect of different sampling strategies on the optimization results and efficiency.

4.1.2 Complexity analysis of the RAST* algorithm

Let n be the total number of iterations of the RAST* algorithm, and the main loop of the RAST* algorithm contains the iterations of the nearest tree node search process and the heuristic search process. The number of iterations of the Nearest() function in the nearest tree node search process is the number of iterations available so far. The number of iterations of the Near() function in the information heuristic search process is also the number of iterations available so far. Then, the time complexity of the processes is represented by the Nearest() function, so the time complexity of the RAST* algorithm and its deformation algorithm is O(n), the time complexity is $O(n^2)$. However, according to existing studies in the literature[35], it is shown that the time complexity of the processes represented by the Nearest() function and the Near() function processes can be reduced to O(logn) by certain methods, i.e., by reducing the number of loops. The nearest tree node position and the set of neighbors can be solved accurately while reducing the number of cycles. Therefore, the RAST* algorithm also can reduce the time complexity to O(nlogn).

The space complexity of the RAST* algorithm is defined as the amount of memory in the storage space for the adaptive sampling tree Tree=(Vertex, Edge), i.e., the size of the Tree set, i.e., Size(Vertex) + Size(Edge). In this paper, the size of the Edge set does not exceed the total number of iterations of the RAST* algorithm, and the size of the Vertex set does not exceed the total number of iterations of the RAST* algorithm plus one because the initialization of the Vertex set already stores the starting point $q_{ts}$. In summary, the RAST* algorithm has a maximum space complexity of O(n+n+1), i.e., O(n).

4.2 Rapidly-exploring information gathering tree algorithm

The RIGT algorithm is a sampling-based motion planning algorithm first proposed by Geoffrey A. Hollinger et al. for the information-driven path planning problem[36]. The advantages of the RIGT algorithm have been analyzed in the literature as it can quickly search the entire workspace and reduce the number of branches and nodes in the tree collection by continuously growing and pruning the tree collection, reducing the number of paths stored in the tree collection. The specific flow of the RIGT algorithm is shown in Algorithm 8.

**Algorithm 8** Rapidly-exploring Information Gathering Tree*

**Enter:** 3D environment model parameters $PA_e$, HAUV related parameters $PA_{hauv}$, task-related parameters $PA_m$, RIGT* algorithm parameters $PA_{rigt}$.

1: Vertex ← {$q_{init}$}; Edge ← ∅; Tree = (Vertex, Edge); $V_{closed}$ ← ∅ $Best_{IG}$ = 0; Bestsol(1) = 0;
2: **for** it = 1 to Max_it **do**
3:     $q_{rand}$ ← RandomSample(IM, $\mathbb{M}$);
4:     $q_{nearest}$ ← Nearest($q_{rand}$, Vertex);
5:     $q_{feasible}$ ← Steer($q_{nearest}$, $q_{rand}$, δ);
6:     $\mathbf{Q_m}$ ← Near(Vertex, $q_{new}$, r);
7:     **for** each $q_m \in Q_m$ **do**
8:         $q_{new}$ ← Steer($q_m$, $q_{feasible}$, δ);
9:         **if** CollisionFree($C_{obs}$, $q_{nearest}$, $q_{new}$) **then**
10:           [$q_{new}$.IG, $q_{new}$.E, $q_{new}$, T, $\mathbb{P}$] = FitnessFun($q_{new}$, $q_m$, $q_{init}$, $q_{final}$, Vertex, $PA_{hauv}$, $PA_e$);
11:           **if** PRUNE($q_{new}$) **then**
12:               Delete $q_{new}$
13:           **else**
14:               Vertex ← Vertex ∪ {$q_{new}$}
15:               Edge ← Edge ∪ {$q_{max}$, $q_{new}$};
16:               **if** $E > E_{max} | T \leq T_{max}$ **then**
17:                   $V_{closed}$ ← $V_{closed}$ ∪ {$q_{new}$}
18:               **else if** $q_{new}$.IG > $Best_{IG}$ **then**
19:                   Best_IG = $q_{new}$.IG
20:               **end if**
21:           **end if**
22:         **end if**
23:     **end for**
24:     Bestsol(it) = Best_IG
25:     **if** Bestsol(it) − Bestsol(it − It_stop) = Best_IG
26:         break:
27:     **end if**
28: **end for**
29: **return** Tree = (Vertex, Edge);

**Output:** $\mathbb{P}^*$, Best_IG, Bestsol

Unlike the RAST* algorithm, the RIGT algorithm is a random sampling strategy, so the parameters of the RIGT algorithm $PA_{rigt}$ include only the step size $\delta$, the neighborhood radius r, the maximum number of iterations Max_it, and the optimization terminated if solution no longer improved after It_stop iterations.

The basic idea of the RIGT algorithm is as follows.

1. Random sampling and nearest point search process: randomly generate sampling points $q_{rand}$ in the workspace and select the nearest sampling point $q_{nearest}$ nearest tree node $q_{rand}$. The branch grows with the length of $\delta$. Then it forms a new node $q_{feasible}$ (Algorithm 8, lines 3-5).

2. Neighborhood access and tree set update process: find the set of neighbors $\mathbf{Q_m}$, for each of the tree nodes in the set of neighbors $q_{feasible}$ grows branches of length up to $\delta$ to the new node $q_{new}$. If neither this branch nor the new node is in the obstacle space and does not need to be cropped, they are put into the sets Edge and Vertex, respectively. Otherwise, delete the new node $q_{new}$; when the new node $q_{new}$ corresponds to a full path whose total energy consumption exceeds the maximum energy or total time consumption exceeds the specified mission time, the new node $q_{new}$ will be placed in the forbidden set $V_{closed}$. The tree nodes in this set no longer grow branches (Algorithm 8, lines 6-17).

3. Update the optimal solution process: update the optimal solution Best_IG and the array of optimal solutions generated after each iteration Bestsol (Algorithm 8, lines 18-27).

4. Output results: After the main loop terminates, the optimal path $\mathbb{p}^*$, the optimal solution $Best\_IG$, and the array of optimal solutions $Bestsol$ generated by each iteration is output.

The rules for whether the newly generated node $q_{new}$ by the RIGT algorithm in step 3 is cropped are defined in the literature as follows[36]. A node $q_{new}$ and its associated node $q_m$, if $q_{new}.IG < q_m.IG, q_{new}.E > q_m.E, q_{new}.T > q_m.T$, the nodes $q_{new}$ are trimmed.

4.3 PSO algorithm

The core idea of the PSO algorithm is as follows, K particles are randomly generated as populations at initialization, and each particle represents a feasible solution. Let $p_k$ and $v_k$ be the position and velocity of the k-th particle, respectively, and the PSO algorithm satisfies the following velocity and position update equations for the kth particle at the i-th iteration.

$$v_k^{i+1} = w^i \cdot v_k^i + c1 \cdot Rand_1^i \cdot (p_{pbest_k}^i - p_k^i) + c2 \cdot Rand_2^i \cdot (p_{gbest}^i - p_k^i) \tag{24}$$

$$p_k^{i+1} = p_k^i \cdot v_k^{i+1} \tag{25}$$

$w^i$ is the weight parameter at the itch iteration, c1 and c2 are the learning factors. $Rand_1^i$, $Rand_2^i$ is [0,1] random number in the interval, $p_{pbest_k}^i$ is the kth particle optimal position at the i-th iteration, $p_{gbest}^i$ is the population optimal position at the i-th iteration. It should be noted here that the weight parameter decays with an increasing number of iterations, satisfying the following equation:

$$w^{i+1} = w^i \cdot w_{damp} \tag{26}$$

Where, $w_{damp}$ is the decay rate of the weight parameter for each iteration. In the PSO algorithm, it is necessary to limit the maximum velocity of each particle in x, y, and z directions $v_{pso}^{max}$, to avoid too large a step; also, it is necessary to constrain that the updated position of each particle cannot overflow the workspace.

---

**Algorithm 9** PSO algorithm

**Enter:** Environment model parameters $PA_e$, HAUV related parameters $PA_{hauv}$, task-related parameters $PA_m$, PSO algorithm parameters $PA_{pso}$.

1: Initialize the position $p^0$ and velocity $v^0$ of each particle to ensure that each particle generates a feasible solution $IG_{pbest}^0$.

2：$IG_{gbest} = \max(IG_{gbest}^0)$;

3：**for** $i = 1$ to Max_it **do**

4：　**for** $k = 1$ to K **do**

5　　Solving for $N_{pso}$ control points in particle k according to Equation. (24) and (25), the $v_k^i$ and $p_k^i$;

6:　　The path nodes from the starting point through the control points to the endpoint are fitted with a B-spline curve to form path $\mathbb{P}$

7:　　　　$IG_k^i = f_I(\mathbb{P})$;

8:　　　　$E_k^i = f_{Energy}(\mathbb{P})$;

9:　　　　$T_k^i = f_{Time}(\mathbb{P})$

10:　　　**if** $IG_k^i > IG_{pbest\_k}^{i-1} \& E_k^i > E_{max} \& T_k^i > T_{max}$**then**;

11:　　　　　$p_{gbest}^i = p_k^i$; $IG_{pbest}^i = IG_k^i$;

12:　　　**else**

13:　　　　　$p_{gbest}^i = p_{gbest}^{i-1}$; $IG_{pbest}^i = IG_k^i$;

14:　　　**end if**

15:　　　**if** $IG_{pbest}^i = IG_{gbest}$**then**

16:　　　　　$p_{gbest}^i = p_{gbest\_k}^i$; $IG_{gbest} = IG_{pbest}^{i-1}$;

17:　　**else**

18:　　　　　$p_{gbest}^i = p_{gbest}^{i-1}$

| | |
|---|---|
| 19: | **end if** |
| 20: | **end for** |
| 21: | $w^i = w^{i-1} \cdot w_{damp}$ |
| 22: | $Bestsol(i) = IG_{gbest}$ |
| 23: | **if** $Bestsol(i) - Bestsol(i - It_{stop}) = 0$ **then** |
| 24: | break: |
| 25: | **end if** |
| 26: **end for** | |
| **Output**: $\mathbb{P}^*, IG_{gbest}, Bestsol$ | |

The PSO algorithm flow is shown in Algorithm 9, and the required input PSO algorithm parameters $PA_{pso}$ include the learning factors c1, c2, initial weight coefficients $w^0$, the weight decay rate $w_{damp}$, maximum particle velocity $v_{pso}^{max}$ population size K, number of control points $N_{pso}$, the maximum number of iterations It_stop, the number of iterations the result is no longer improved Max_it. The PSO algorithm updates the position and velocity of each control point according to Equations (24) and (25).

5 SIMULATION EXPERIMENT RESULTS AND ANALYSIS

In this section, the RAST*-I/E algorithm, RAST*-I algorithm, RAST algorithm, RRST* algorithm, RRST algorithm and PSO algorithm designed in subsection four are compared in five simulation cases under various scenarios.

5.1 Simulation experiment setup

All simulation experiments in this section were performed on a host computer with Windows 10 operating system, Intel(R) Core(TM) i7-6700HQ CPU @ 3.40 GHz and 16.0 GB of RAM. The parameter settings of the HAUV and all algorithms in the simulation experiments are shown in Table 2. The parameters of the HAUV in this paper are based on the values of the HAUV "Nezha"[51]. The optimal power of the HAUV for air flight $P_{air}$, the optimal power for underwater navigation $P_{sea}$ and single energy consumption for cross-media motion mode $E_{switch\_1}$, all three parameters are related to the HAUV with finite energy $E_{max}$. In this paper, $E_{max}$ is used as a criterion to invert $P_{air}$, $P_{sea}$ and $E_{switch\_1}$ relative to the scale factor of $E_{max}$, thus determining the relative values of these three parameters. In addition, the basic parameter settings of all algorithms in this paper are based on the summary of existing research literature. The step size $\sigma$ and the neighborhood radius r of the RAST* algorithm are based on the size of the environmental raster map. In particular, it should be noted that the results of this paper set It_stop is set to 200 times to speed up the solution of the algorithm.

This section focuses on the simulation experiments of the information-driven path planning problem for a single HAUV. The following five scenarios are designed.

Scenario 1: Path planning for a HAUV with limited energy.

Scenario 2: Path planning for a HAUV under the dual constraints of limited energy and mission time.

Scenario 3: Path planning for a HAUV under dual constraints of limited energy and tight mission time.

Scenario 4: Path planning for a HAUV with a higher weight of information on ocean features than on atmosphere.

Scenario 5: Path planning for a HAUV with higher weight of information on the atmospheric than the ocean.

The air and underwater environmental information weights are the same in Scenarios 1-3, and these simulations mainly demonstrate the performance of these six algorithms under different constraints. In Scenarios 4-5, different weights are set in the information collection function for the aerial and underwater features, and the simulations compare the optimization performance of these six algorithms.

TABLE 2 Parameter settings for the HAUV and all algorithms

| | parameters | notation | value |
|---|---|---|---|
| HAUV | air speed | $V_{air}$ | 10(m/s) |
| | power for air flight | $P_{air}$ | $\frac{1}{900}E_{max}$ |
| | underwater speed | $V_{sea}$ | 0.5(m/s) |
| | power for underwater navigation | $P_{sea}$ | $\frac{1}{28800}E_{max}$ |
| | power consumption in cross-media | $E_{switch\_1}$ | $\frac{1}{30}E_{max}$ |
| | time consumption for cross-media | $T_{switch\_1}$ | 20(s) |
| | limited energy | $E_{max}$ | One standard unit |
| | sensor perception factor | $\mathcal{A}_{d_{max}}$ | 1 |
| | sensor distance attenuation coefficient | $\sigma$ | 1 |
| | sensing range | $d_{max}$ | 100(m) |
| RAST*Algorithm | Number of tournament selection | $\mathbb{M}$ | 10 |
| | Step length | $\delta$ | 5 |
| | Neighborhood radius | $r$ | 10 |
| | Maximum number of iterations | $V_{sea}$ | 5000 |
| | Number of times the result is no longer improved | It_stop | 200 |
| PSO Algorithm | Learning factor | c1 | 1 |
| | Learning factor | c2 | 1 |
| | Initial weighting factor | $\omega^0$ | 1 |
| | Decay rate of weights | $\omega_{damp}$ | 0.99 |
| | Maximum particle velocity | $v_{pso}^{max}$ | [5,5.1] |
| | Population size | $K$ | 50 |
| | Number of control points | $N_{pso}$ | 5 |

The environmental model used in the above five scenarios is a 100×100×13 raster map, representing a mission area of 5km×5km×600m. Each raster point in the raster map contains environmental feature information and velocity field data. The distribution of the values of all raster points in the workspace is [0,1]. To evaluate the algorithm's performance, several runs of experiments are performed for each algorithm, and the following performance indicators are introduced to measure and evaluate the strengths and weaknesses of the algorithm in terms of computational accuracy and efficiency.

(1) The average information collection $I_{mean}$, $I_{mean}$ satisfies the following equation:

$$I_{mean} = \frac{1}{\mathbb{N}}\sum_{i=1}^{\mathbb{N}} I_i \qquad (27)$$

$\mathbb{N}$ is the number of samples, i.e., the times of experiment replications.

(2) The standard deviation of the information collection $I_{std}$, $I_{std}$ satisfies the following equation.

$$I_{std} = \sqrt{\frac{1}{\mathbb{N}-1}\sum_{i=1}^{\mathbb{N}}|I_i - I_{mean}|^2} \qquad (28)$$

(3) The average number of iterations.

(4) The average computation time.

5.2 Information-driven path planning for HAUV under different constraints

In this subsection, we analyze the optimization performance of the above six algorithms to solve the HAUV information-driven path planning problem under different constraints when the information weights of the sea and air environment features are the same.

5.2.1 Scenario 1: Path planning for a HAUV with limited energy

First, consider a HAUV carrying a finite energy $E_{max}$, assume that the starting position of the HAUV is $q_{init}$= (1km,3.75km,0m) and the mission end position is $q_{final}$=(4km,3.75km,0m) and the mission time is set to $T_{max}$=inf, the mission area is based on the analytic equation, and there exists a submerged obstacle space similar to the continental slope. In this parameter setting, each algorithm is repeated ten times to verify the algorithms' stability, average optimization performance, and average computational speed.

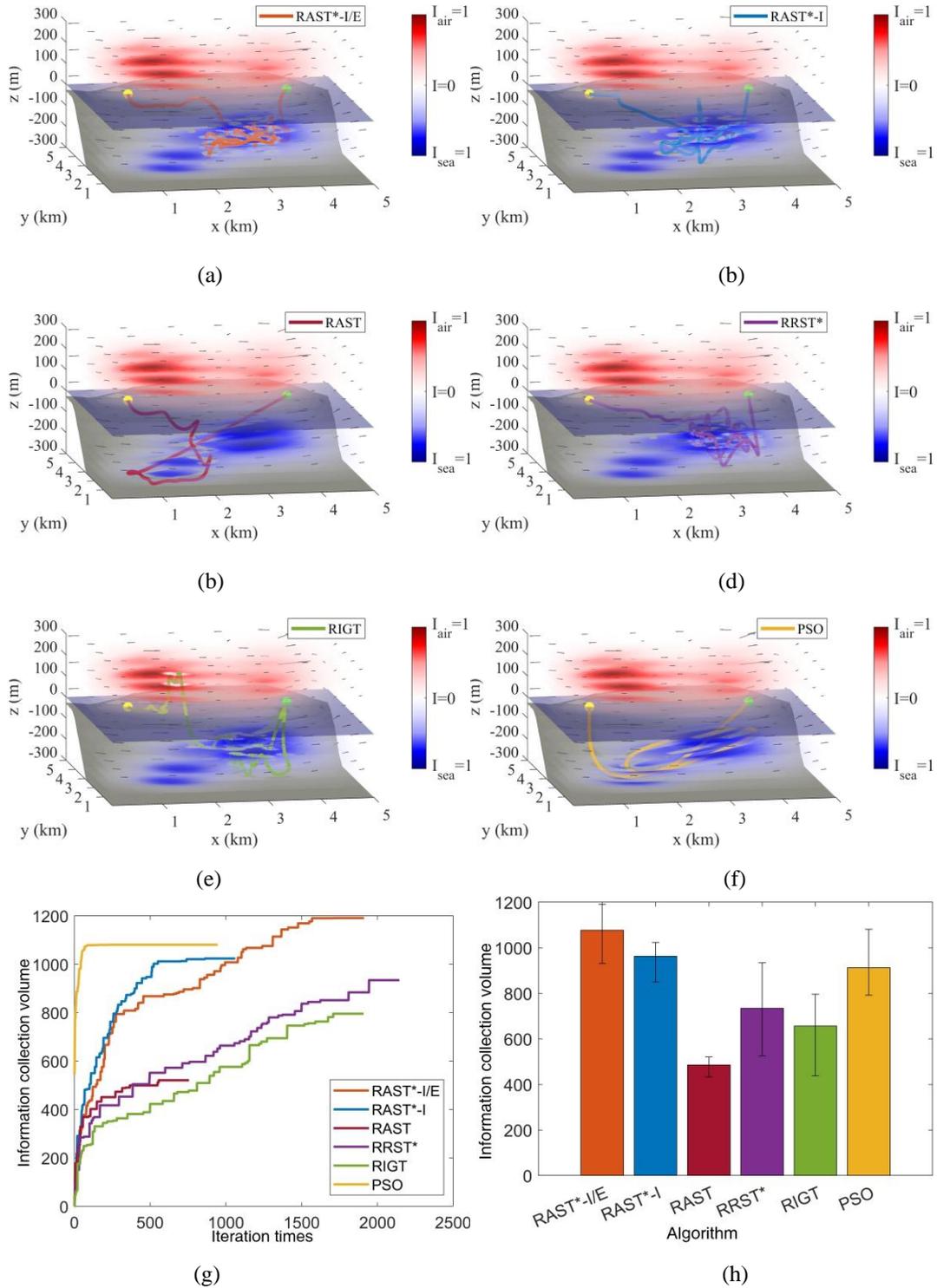

Figure 5 Scenario 1: Informative path, convergence curve and error bar produced by the path planners

TABLE 3 Scenario 1: Comparison of the optimal results of the simulation experiment algorithm

| Algorithm | Amount of information collected | Iteration times | Energy consumption ($E_{max}$) | Task execution time (h) |
|---|---|---|---|---|
| RAST*-I/E | 1190.87 | 1912 | 0.99 | 7.41 |
| RAST*-I | 1023.72 | 1062 | 1.00 | 6.56 |
| RAST | 520.96 | 757 | 0.71 | 5.17 |
| RRST* | 934.25 | 2145 | 0.99 | 6.97 |
| RIGT | 795.79 | 1909 | 0.99 | 4.58 |
| PSO | 1080.51 | 946 | 1.00 | 7.48 |

TABLE 4 Scenario 1: Algorithm performance comparison

| Algorithm | Average message size | Standard deviation | Average number of iterations | Average computation time (s) |
|---|---|---|---|---|
| RAST*-I/E | 1077.20 | **84.01** | 1401 | 52 |
| RAST*-I | 963.08 | **58.07** | 951 | 30 |
| RAST | 485.21 | **28.79** | 611 | 3 |
| RRST* | 734.70 | **171.23** | 1294 | 43 |
| RIGT | 656.06 | **121.73** | 1360 | 34 |
| PSO | 912.47 | **108.99** | 828 | 46 |

Figure 5 show the HAUV sampling path, the distribution and convergence curve of environmental feature information collected along this path, and the error plot of information collection for ten repetitions of each algorithm. In the figure, the obstacle space is represented by a grey surface, the starting point is a yellow dot, and the endpoint is a green dot. The RAST*-I/E algorithm, RAST*-I algorithm, RAST algorithm, RRST*, RIGT, and PSO algorithms are orange, blue, red-brown, purple, lime green, and yellow-brown, respectively. From the graph of the optimized sampling paths of each algorithm, it can be seen that there is no limit on the RIGT algorithm performing aerial sampling in the pre-task period. The main reason is that the sampling volume per unit energy consumption of the HAUV for underwater sampling is much higher than that for aerial sampling. In the absence of a mission time constraint, a global algorithm would choose to slowly collect the amount of information in the environment for a longer period to save the energy of the HAUV. However, the RIGT algorithm is relatively weak in global optimization, and the algorithm does not learn the experience of increasing the amount of information collected by extending the sampling time through iterations. Hence, the optimized path information collection is less.

The convergence curves in Figure 5g show that the RAST*-I/E algorithm has the most information collection. The three algorithms with relatively slow convergence are the RAST*-I/E algorithm, the RRST* algorithm, and the RIGT algorithm, which require a relatively large number of iterations to find the asymptotically optimal solution. Figure 5h shows the deviation of the results for each algorithm for ten repeated experiments. The two algorithms with larger deviation are RRST* algorithm and RIGT algorithm, which can also be seen from the standard deviation in Table 4. The reason may be that these two algorithms' sampling strategies are random, lacking good environmental information to guide the algorithm, which may repeatedly search near a local solution so that the results are no longer improved in a certain iterative process.

From the results recorded in Table 3, we can find that, except for the RAST algorithm, the optimized paths of the other five algorithms almost exhaust the energy, meaning energy usage is maximized. The main reason is that, although the tournament-based point selection method can guide the adaptive sampling tree to grow toward regions with high values, the algorithm structure does not have the process of initiation search and parent node reshaping. The algorithm lacks initiation and does not re-evaluate the adaptability of the tree nodes to the newly generated nodes. The simulation results show that the computation time of the RAST algorithm is significantly better than the rest of the algorithms. Still, it can only solve the feasible path, not the global optimal path, and thus the RAST algorithm has the lowest optimization performance.

From Tables 3 and 4, we can also obtain that the three algorithms with the best optimization performance are the RAST*-I/E algorithm, the RAST*-I algorithm and the PSO algorithm. All three algorithms have an optimal information collection of more than 1000. However, it can also be seen in Figure 5 that the paths generated by the RAST*-I/E algorithm are sampled back and forth underwater. In contrast, the paths generated by the PSO algorithm generates paths with a shape of S, and ultimately the PSO algorithm does not collect as much information as the RAST*-I/E algorithm.

Comparing the simulation results of RAST*-I/E algorithm and the RAST*-I algorithm, the RAST*-I/E algorithm has slightly higher optimization performance than the RAST*-I algorithm but slightly lower optimization speed. The only difference between these two algorithms is the heuristic factor in the information heuristic search process. The heuristic factor of the RAST*-I/E algorithm adopts the information collection amount per unit energy consumption of the path. In contrast, the heuristic factor of the RAST*-I algorithm adopts the information collection amount directly, which means that the information heuristic search process of the RAST*-I algorithm does not consider the power consumption of the HAUV. On the other hand, the RAST*-I/E algorithm considers the power consumption of the HAUV, and the iterative search process is oriented toward collecting more information with less energy, but this increases the number of iterations. Therefore, the RAST*-I/E algorithm is not as fast as the RAST*-I algorithm, but its optimization performance is better.

5.2.2 Algorithm 2: Path planning for a HAUV under the constraints of limited energy and mission time

In a real mission, marine scientists usually specify the mission time so that information can be obtained on time, which also facilitates the recovery of the HAUV. The environmental characteristics of the target area are downloaded from the official NOAA website for a small area in the Gulf of Mexico. Since the raster scale of the data provided by the official NOAA website is too large, this section scales the dataset to store the same feature information in a square environment of 50m in length, width and height for each raster. The dataset is normalized to $I_{air}, I_{sea} \in [0,1]$. Set the HAUV's starting position as $q_{init}=$ (0.5km,2.5km,0m) and the mission end position as $q_{init}=$ (4.5km,2.5km,0m), the mission time is $T_{max}=$ 3h.

From Figure 6, it can be seen that the RAST*-I/E algorithm has the least remaining red and blue areas compared to other algorithms. It means that the RAST*-I/E algorithm captures the most information, which is quantitatively shown in Figure 6g and Table 5. Combining Figure 6h and Table 6, it can be found that the three algorithms with the best optimization performance are still the RAST*-I/E algorithm, RAST*-I algorithm and PSO algorithm.

The optimal results in Table 5 show that the path generated by the RAST*-I/E algorithm and the RAST*-I algorithm used almost all the power and mission time. The RRST*, RIGT and PSO algorithms reach the boundary of one single constraint and have a fair search capability, while the RAST algorithm still only generates feasible solutions. The performance metrics in Table 6 still reflect a similar situation to Scenario 1, i.e., the random sampling strategy makes the RRST* algorithm and the RIGT algorithm require more iterations to find the global solution. The RAST*-I/E algorithm performs better under multiple constraints than one single constraint.

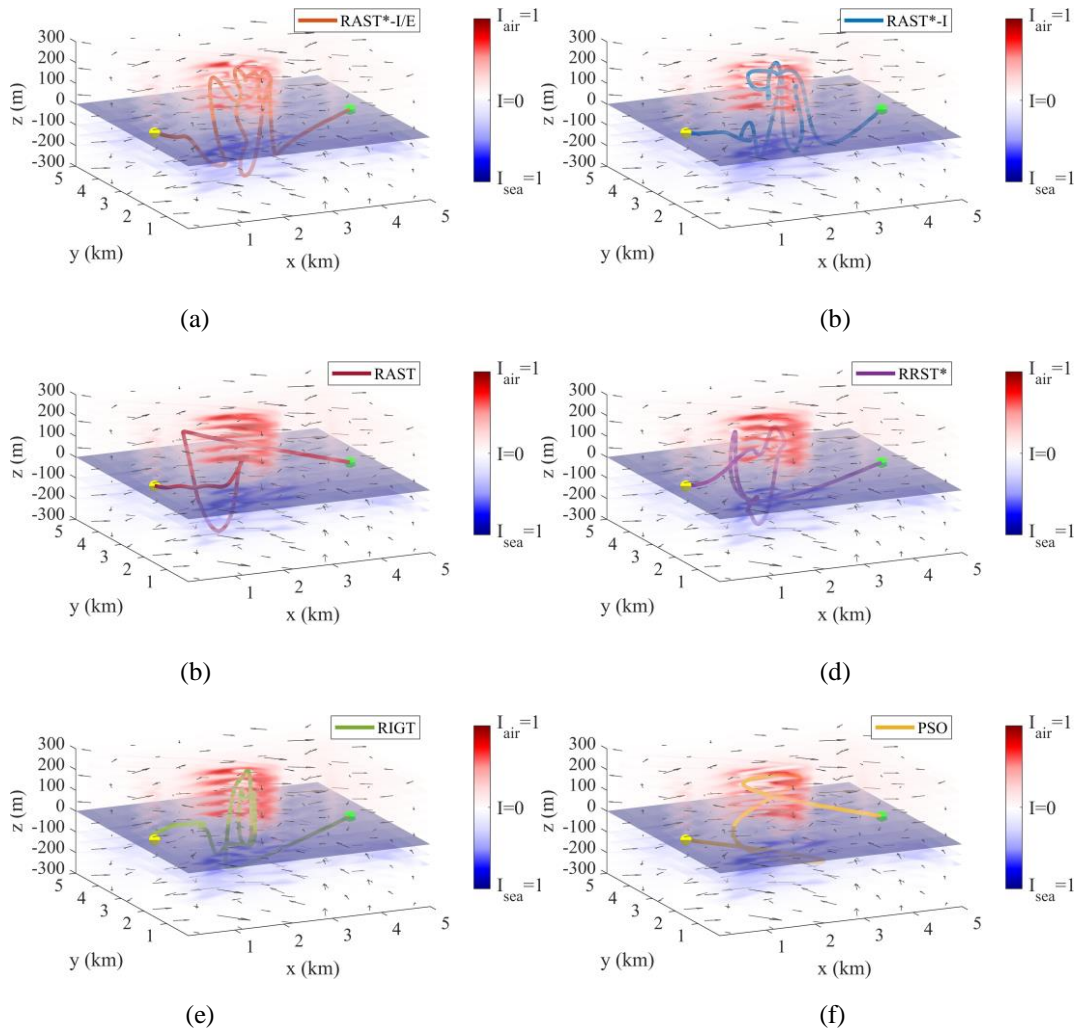

(a)

(b)

(b)

(d)

(e)

(f)

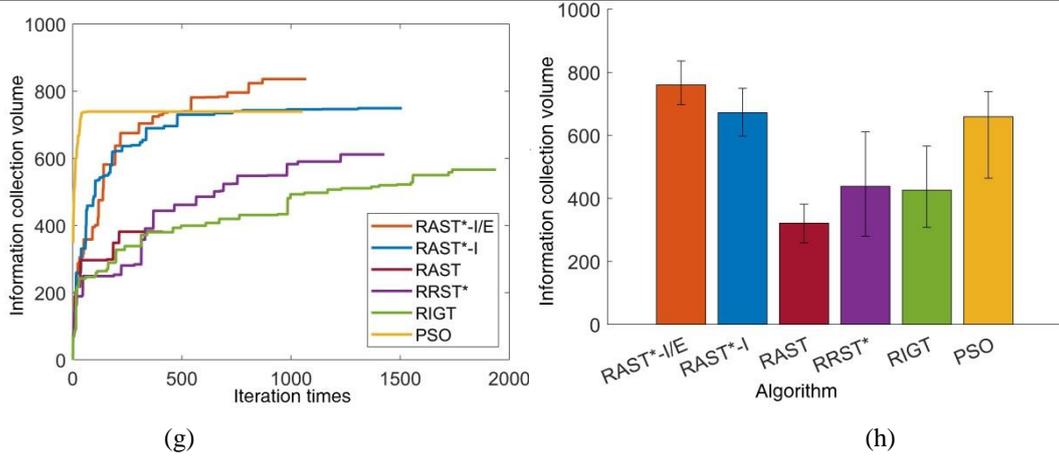

(g)　　　　　　　　　　　　　　　　　　　　(h)

Figure 6 Scenario 2: Informative path, convergence curve and error bar produced by the path planners

TABLE 5 Scenario 2: Comparison of the optimal results of the simulation experiment algorithm

| Algorithm | Amount of information collected | Iteration times | Energy consumption ($E_{max}$) | Task execution time (h) |
|---|---|---|---|---|
| RAST*-I/E | 836.08 | 1070 | 0.99 | 2.99 |
| RAST*-I | 749.42 | 1508 | 1.00 | 2.99 |
| RAST | 381.63 | 413 | 0.88 | 1.46 |
| RRST* | 611.68 | 1428 | 0.89 | 2.97 |
| RIGT | 566.38 | 1939 | 0.97 | 2.89 |
| PSO | 739.35 | 1054 | 1.00 | 2.88 |

TABLE 6 Scenario 2: Repeated experimental algorithm performance metrics comparison

| Algorithm | Average message size | Standard deviation | Average number of iterations | Average computation time (s) |
|---|---|---|---|---|
| RAST*-I/E | 760.21 | 44.73 | 760 | 41 |
| RAST*-I | 672.20 | 59.80 | 766 | 43 |
| RAST | 321.70 | 39.59 | 549 | 3 |
| RRST* | 438.34 | 132.26 | 837 | 21 |
| RIGT | 426.40 | 74.17 | 977 | 25 |
| PSO | 659.49 | 84.57 | 616 | 45 |

5.2.3 Scenario 3: Path planning for a HAUV under the constraints of limited energy and tight mission time

While the mission time studied in Scenario 2 is more relaxed, this scenario will conduct experiments for path planning of a HAUV under a tighter mission time. It is assumed that the HAUV needs to be recovered at this location after one hour. As shown in Figure 7, the areas with high information values of the atmosphere are distributed in the range of [3km, 5km] on the x-axis. The areas with high information values of oceanic features are distributed in the range of [1km, 3km] on the x-axis. Due to the tight mission time, the HAUV prefers to collect atmospheric feature information, but it is constrained by the limited energy to perform only aerial sampling. The optimized path maximizes atmospheric and oceanic information collection under double constraints. The convergence curves in Figure 7g show that the RRST* algorithm based on the random sampling strategy and the RIGT algorithm has the highest convergence. The RIGT algorithm converges the slowest. The information acquisition error in Figure 7h shows that the optimization capability of the algorithms in this example can be roughly divided into three echelons. The first echelon is the RAST*-I/E algorithm, the second echelon is the RAST*-I algorithm, RRST* algorithm and PSO algorithm, and the third echelon is the RAST algorithm and RIGT algorithm.

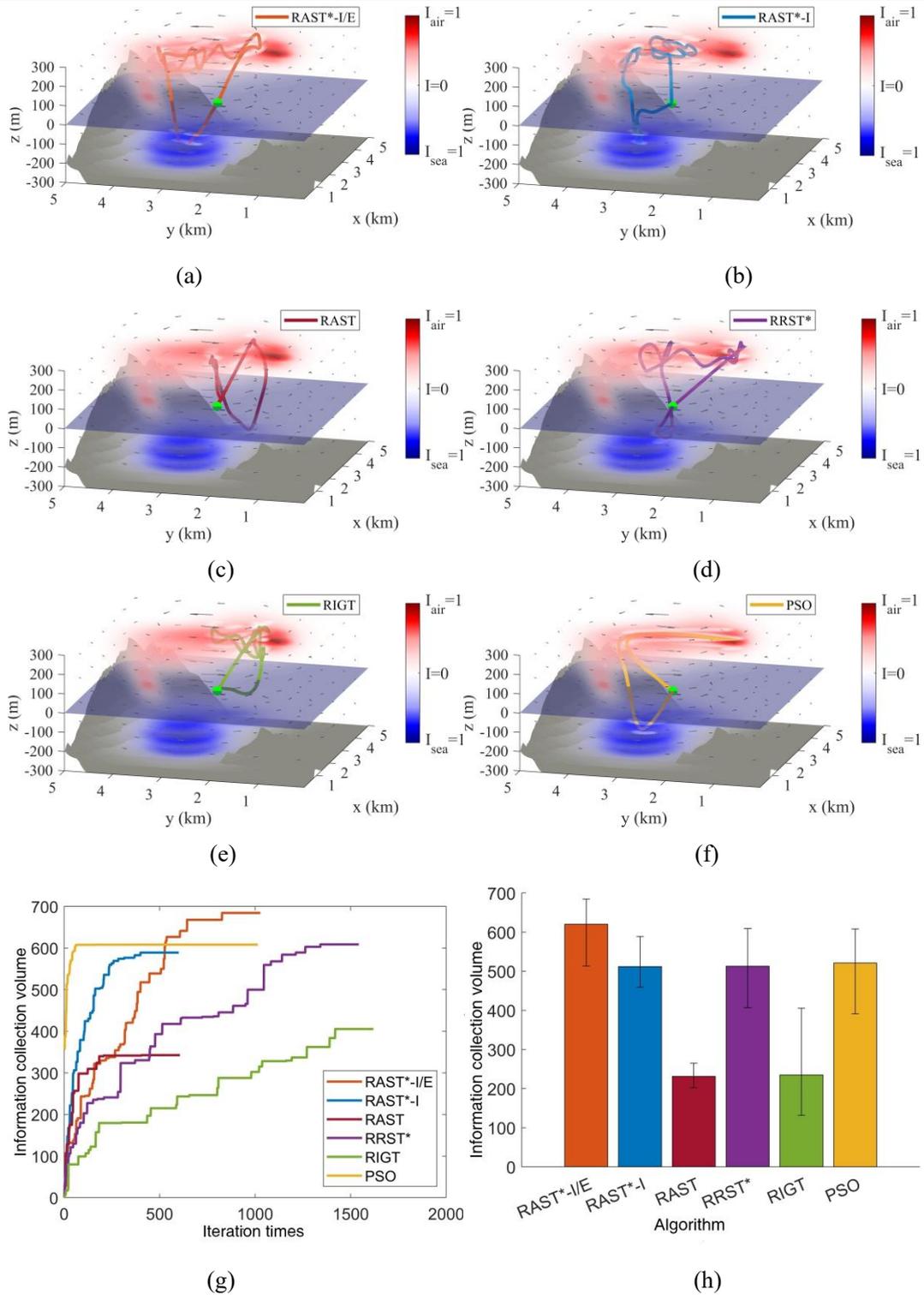

Figure 7 Scenario 3: Informative path, convergence curve and error bar produced by the path planners

TABLE 7 Scenario 3: Comparison of the optimal results of the simulation experiment algorithm

| Algorithm | Amount of information collected | Iteration times | Energy consumption ($E_{max}$) | Task execution time (h) |
|---|---|---|---|---|
| RAST*-I/E | 684.52 | 1027 | 1.00 | 0.94 |
| RAST*-I | 589.14 | 600 | 0.99 | 0.99 |
| RAST | 342.72 | 606 | 0.72 | 0.74 |
| RRST* | 608.90 | 1542 | 0.99 | 0.94 |
| RIGT | 405.55 | 1619 | 0.96 | 0.89 |
| PSO | 608.15 | 1014 | 0.99 | 1.00 |

TABLE 8 Scenario 3: Repeated experimental algorithm performance metrics comparison

| Algorithm | Average message size | Standard deviation | Average number of iterations | Average computation time (s) |
|---|---|---|---|---|
| RAST*-I/E | 620.33 | 48.60 | 999 | 45 |
| RAST*-I | 511.98 | 41.10 | 831 | 32 |
| RAST | 231.18 | 18.43 | 564 | 2 |
| RRST* | 513.00 | 63.02 | 1351 | 41 |
| RIGT | 234.90 | 87.40 | 695 | 26 |
| PSO | 521.01 | 60.59 | 750 | 63 |

From Table 7, we can calculate that the optimal result of the RAST*-I/E algorithm has 12% higher information collection than the RRST* algorithm. The reason is that the RAST*-I/E algorithm introduces the tournament point selection method into the sampling strategy. Therefore, more sampling points fall in the regions where feature information is gathered. From Table 8, it can be calculated that the average information collection of the optimized results of the RAST*-I/E algorithm is 19% higher than that of the second. The simulation results of this example fully verify the optimization performance and stability of the RAST*-I/E. However, the speed of the RAST*-I/E algorithm is only moderate. In addition, the average computation time of the PSO algorithm in Table 8 is higher than that of the remaining five algorithms, mainly because the process of finding feasible solutions for the initial particle swarm is uncertain.

5.3 Information-driven path planning for a HAUV with weighted environmental feature information

This subsection analytically investigates the performance of the above six algorithms for solving the HAUV in different weights of information on the sea and air environment.

5.3.1 Algorithm 4: Path planning for a HAUV with higher weight of ocean feature information than the atmosphere

It is assumed that marine scientists are more interested in information of the ocean, assigning a weighting factor of ocean feature information weighting factor $\kappa_{sea}=3$ and the weighting factor of atmospheric feature information $\kappa_{air}=1$.

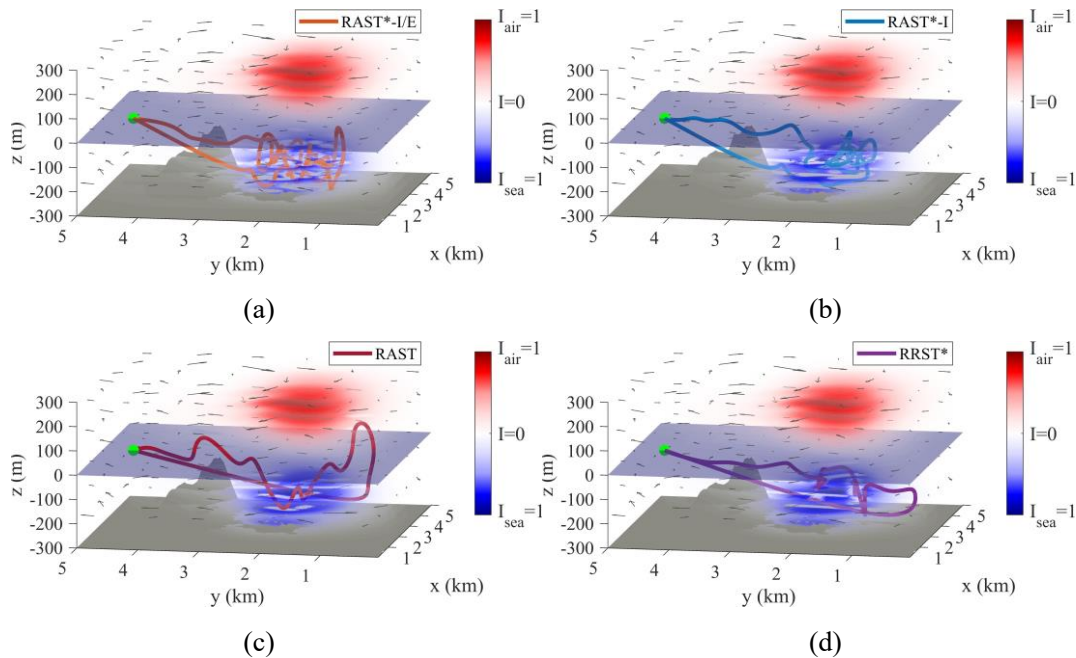

(a) (b)
(c) (d)

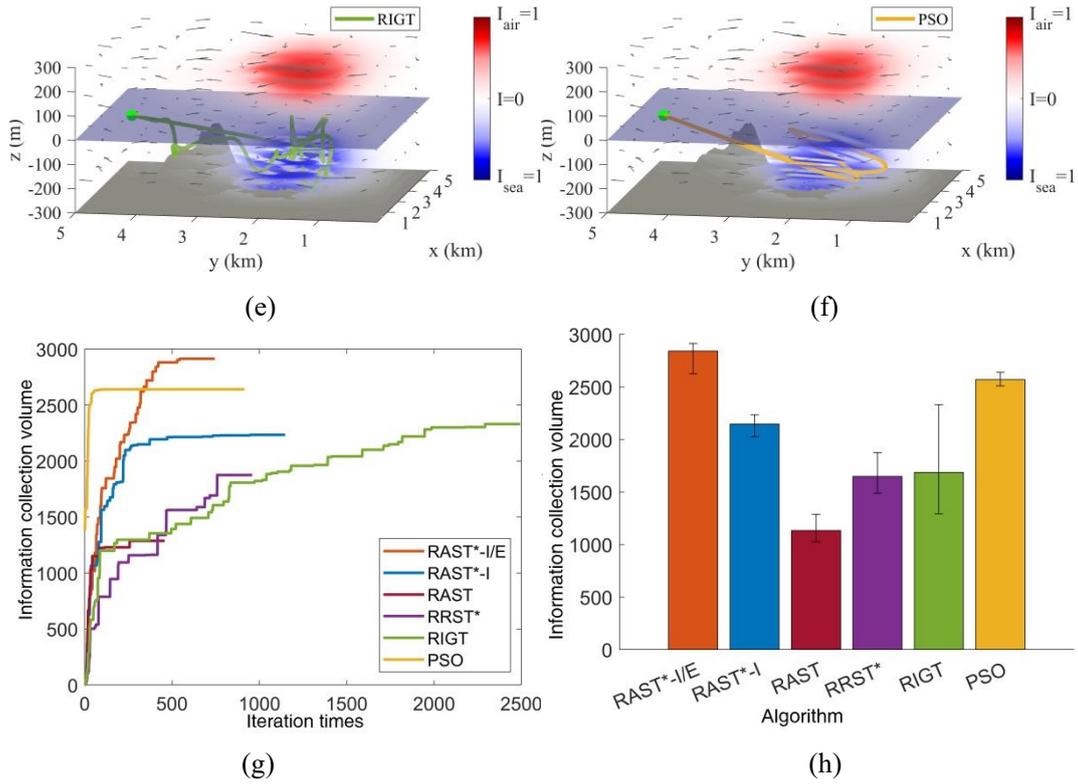

Figure 8 Scenario 4: Informative path, convergence curve and error bar produced by the path planners

TABLE 9 Scenario 4: Comparison of the optimal results of the simulation experiment algorithm

| Algorithm | Amount of information collected | Iteration times | Energy consumption ($E_{max}$) | Task execution time (h) |
|---|---|---|---|---|
| RAST*-I/E | 2914.08 | 745 | 1.00 | 7.47 |
| RAST*-I | 2234.58 | 1148 | 1.00 | 6.21 |
| RAST | 1288.18 | 458 | 0.94 | 4.09 |
| RRST* | 1875.60 | 959 | 0.90 | 6.67 |
| RIGT | 2331.55 | 2493 | 1.00 | 6.89 |
| PSO | 2641.41 | 915 | 1.00 | 7.48 |

TABLE 10 Scenario 4: Repeated experimental algorithm performance metrics comparison

| Algorithm | Average message size | Standard deviation | Average number of iterations | Average computation time (s) |
|---|---|---|---|---|
| RAST*-I/E | 2841.31 | 87.36 | 1008 | 54 |
| RAST*-I | 2147.38 | 74.94 | 866 | 46 |
| RAST | 1132.68 | 86.04 | 528 | 2 |
| RRST* | 1648.89 | 139.46 | 987 | 45 |
| RIGT | 1686.39 | 416.07 | 1532 | 69 |
| PSO | 2569.83 | 40.51 | 947 | 94 |

As shown in Figure 8, each algorithm crosses the obstacle space and travels to the region where environmental feature information is gathered. Due to the high weight of ocean information, algorithms with high optimization capabilities focus on collecting underwater information. This conclusion is corroborated in figure 8, where only the RAST algorithm activates the cross-media motion mode for the HAUV. However, the optimized path does not capture more environmental information, so the RAST algorithm is still not able to search for the global solution. From Figure 8g and Figure 8h, it can be seen that the optimization speed and stability of the RIGT algorithm is poor. From Table 9, we can find that the RAST*-I/E algorithm, RAST*-I algorithm, RIGT algorithm, and PSO algorithm all make full use of the limited power of the HAUV before returning to the base. Hence, the optimal results of all four algorithms exceed 2000. From Table 10, the three algorithms with the best optimization performance are still RAST*-I/E, RAST*-I, and PSO. The optimization performance of the RAST*-I/E algorithm is 10.6% higher than that of the PSO algorithm.

5.3.2 Scenario 5: Path planning for a HAUV with a higher weighting of atmospheric feature information than the ocean

Assuming that marine scientists are more interested in atmospheric information, a weighting factor $\kappa_{air}=3$ is assigned to atmospheric information and a weighting factor $\kappa_{sea}=1$ to oceanic information.

From the paths of each algorithm shown in Figure 9, it can be found that, except for the RAST algorithm, all the other five algorithms only perform aerial sampling before heading to the endpoint, which is due to the high weight of atmospheric feature information. According to the convergence curves Figure 9g, it is found that the PSO algorithm and RAST*-I/E algorithm generate the path with maximum information collection. In Tables 11 and 12, the optimal results and the performance indexes show that both algorithms perform well in repeated experiments. The optimization performance of the PSO algorithm is slightly higher than that of the RAST*-I/E algorithm, and the number of iterations is reduced by nearly two times, which highlights that the PSO algorithm is indeed an excellent global optimization algorithm. However, the standard deviation of the PSO algorithm in repeated runs is higher than that of the RAST*-I/E algorithm, indicating that the PSO algorithm is not as stable as the RAST*-I/E algorithm. Moreover, the average computation time of the PSO algorithm is slightly longer than that of the RAST*-I/E algorithm, indicating that a large amount of time is wasted in finding the initial feasible solution. Still, once the initial feasible solution of the particle swarm is generated, the convergence speed of the algorithm's main loop can be accelerated, thus reducing the number of iterations. In summary, for this scenario, if a high stability algorithm is needed, this paper recommends the RAST*-I/E algorithm; if a fast convergence algorithm is required, this paper recommends the PSO algorithm.

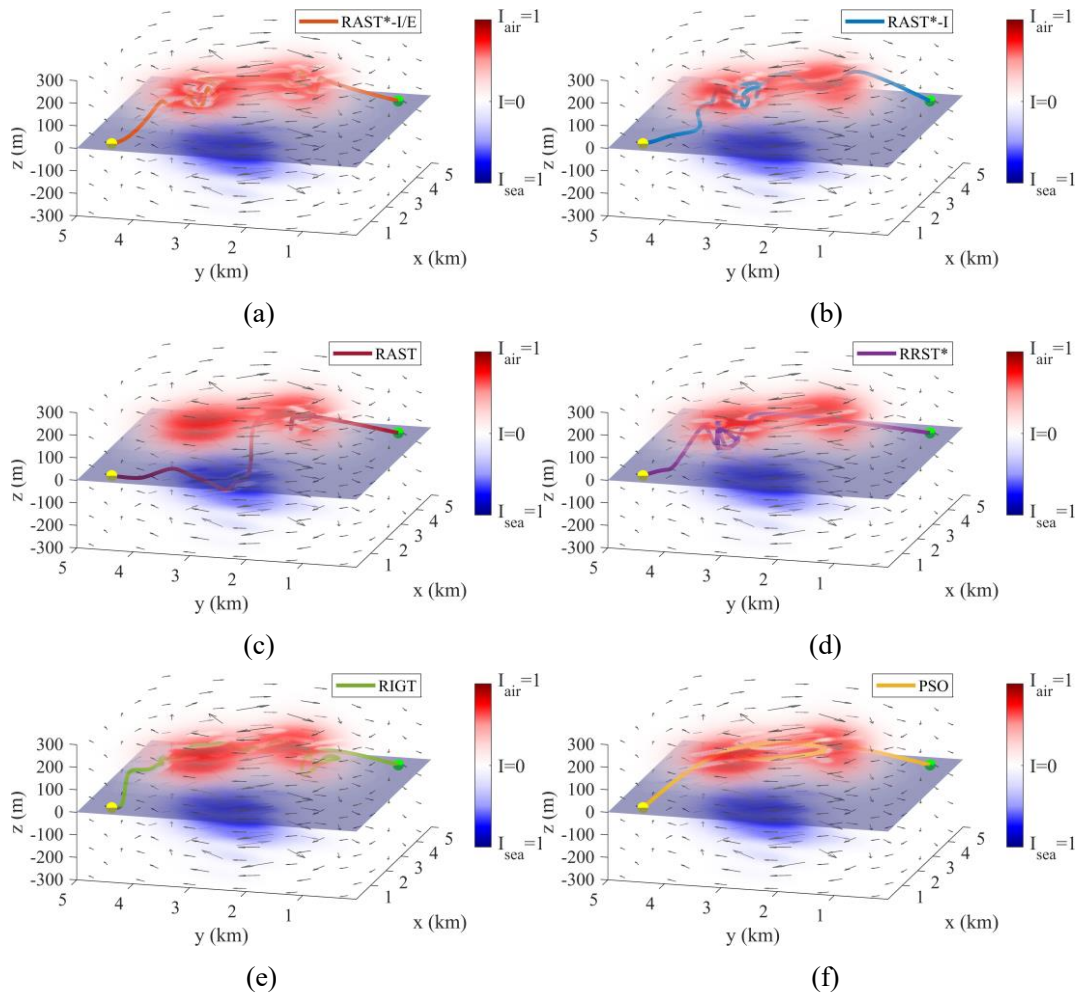

(a)  (b)  (c)  (d)  (e)  (f)

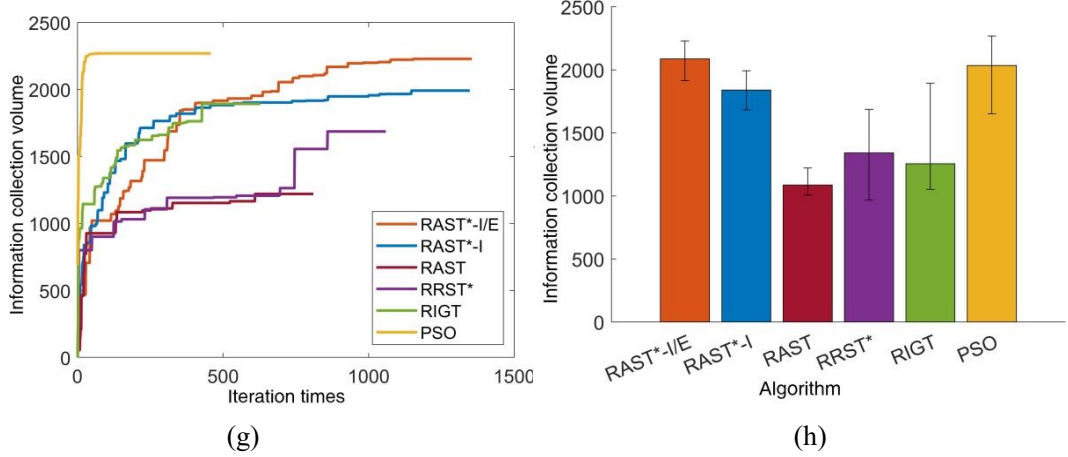

(g)            (h)

Figure 9 Scenario 5: Informative path, convergence curve and error bar produced by the path planners

TABLE 11 Scenario 5: Comparison of the optimal results of the simulation experiment algorithm

| Algorithm | Amount of information collected | Iteration times | Energy consumption ($E_{max}$) | Task execution time (h) |
| --- | --- | --- | --- | --- |
| RAST*-I/E | 2229.48 | 1354 | 1.00 | 0.24 |
| RAST*-I | 1992.40 | 1347 | 1.00 | 0.24 |
| RAST | 1221.39 | 810 | 0.85 | 1.80 |
| RRST* | 1687.20 | 1059 | 0.94 | 0.23 |
| RIGT | 1893.54 | 628 | 1.00 | 0.33 |
| PSO | 2270.10 | 458 | 1.00 | 0.24 |

TABLE 12 Scenario 5: Repeated experimental algorithm performance metrics comparison

| Algorithm | Average message size | Standard deviation | Average number of iterations | Average computation time (s) |
| --- | --- | --- | --- | --- |
| RAST*-I/E | 2088.10 | 98.65 | 1193 | 62 |
| RAST*-I | 1839.98 | 100.79 | 993 | 44 |
| RAST | 1085.83 | 72.46 | 414 | 1 |
| RRST* | 1341.29 | 254.53 | 543 | 23 |
| RIGT | 1255.39 | 249.53 | 607 | 26 |
| PSO | 2034.48 | 191.10 | 583 | 71 |

5.4 Robustness Analysis

To verify the robustness of these six algorithms, this subsection will conduct simulation experiments to evaluate three scenarios from single constraint condition, double constraint condition, and the sea-air environment with different weights, respectively. Each scenario is simulated 100 times, i.e., 100 different maps of sea-air environments and velocity fields are randomly generated, and the task start and end positions are randomly selected. The specific settings of the three mission scenarios are as follows.

- Scenario 1: the HAUV carries a limited energy $E_{max}$, but has no limits for the mission time.
- Scenario 2: the HAUV carries a limited energy $E_{max}$ with a mission time of a random number in the interval [1h, 3h].
- Scenario 3: the HAUV carries a limited energy $E_{max}$ with a mission time of $T_{max}$= 3h and atmospheric and oceanic information weights $\kappa_{air}$ and $\kappa_{sea}$ are random integers in the interval [1,5].

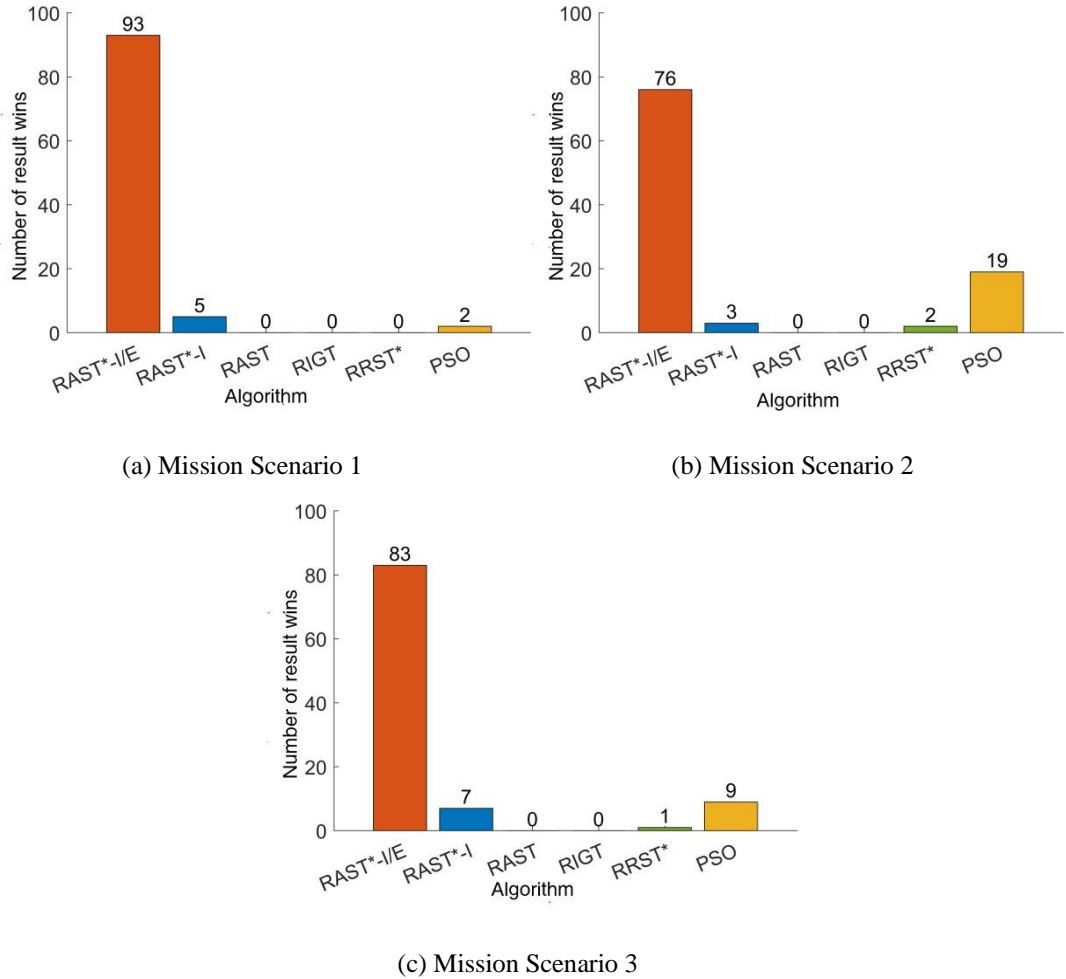

(a) Mission Scenario 1    (b) Mission Scenario 2

(c) Mission Scenario 3

Figure 10 Robustness assessment results

The robustness experiments are judged by which algorithm can collect more information with the same setting. The algorithm with the highest number of winning results among 100 randomized simulations for each scenario is considered the most robust and more suitable for practical applications in such scenarios.

The results are shown in Figs. 10. The RAST*-I/E algorithm wins over the other five algorithms by an absolute margin, indicating that the paths optimized by the RAST*-I/E algorithm are more likely to capture more information, confirming the superiority of the RAST*-I/E algorithm in solving the HAUV information-driven path planning problem. On the other hand, as an excellent classical global optimization algorithm, the PSO algorithm is slightly less robust than the RAST*-I/E algorithm proposed in this paper in terms of adaptability and optimization capability.

6 SUMMARY

This paper presents a new RAST*-I/E algorithm for information-driven path planning problems of HAUV. This RAST*-I/E algorithm innovatively combines the sampling strategy based on the tournament point selection method, information heuristic search processand the framework of RRT* algorithm. In order to compare the effectiveness of the newly designed structure in the RAST*-I/E algorithm, the RAST*-I algorithm with the heuristic factor of total information is designed according to the information heuristic search process; the RRST* algorithm based on the random sampling strategy is designed according to the sampling strategy; and the RAST without the information heuristic search and parent node reshaping process is designed according to the presence or absence of this process algorithm without this process according to the presence or absence of information heuristic search and parent node reshaping process. Moreover, the classical RIGT algorithm and the PSO algorithm are designed as the comparison algorithm. The simulation experiments were conducted to compare the above six algorithms' optimization performance, speed, and stability through five cases. The sampling strategy of the RAST*-I/E algorithm based on the tournament point selection method guides the adaptive sampling tree to explore the regions where information with higher values are located. The information heuristic search process is the key to preventing the algorithm from falling into local optimal paths. The RAST*-I/E algorithm combines the advantages of tournament point selection, information heuristic search, and RRT* algorithm to efficiently search the air-sea environment. Therefore, the obtained sampling path can collect the most information, which ensures the accuracy and robustness of the algorithm.

ACKNOWLEDGMENTS

This Research is supported in part by the National Natural Science Foundation of China under grant 41706108 and in part by the Science and Technology Commission of Shanghai Municipality Project 20dz1206600 and in part by the Natural Science Foundation of Shanghai under Grant 20ZR1424800 and in partly by the Shanghai Jiao Tong University Scientific and Technological Innovation Funds under Grant 2019QYB04.